\documentclass[letterpaper]{article} % DO NOT CHANGE THIS
\usepackage{aaai24}
\usepackage{times}  % DO NOT CHANGE THIS
\usepackage{helvet}  % DO NOT CHANGE THIS
\usepackage{courier}  % DO NOT CHANGE THIS
\usepackage[hyphens]{url}  % DO NOT CHANGE THIS
\usepackage{graphicx} % DO NOT CHANGE THIS
\usepackage{natbib}  % DO NOT CHANGE THIS AND DO NOT ADD ANY OPTIONS TO IT
\usepackage{caption} % DO NOT CHANGE THIS AND DO NOT ADD ANY OPTIONS TO IT
\usepackage{macros}
\usepackage[english]{babel}
\usepackage{amsthm}
\usepackage{graphicx, array, blindtext}
\usepackage{dblfloatfix}
\theoremstyle{definition}
\newtheorem{definition}{Definition}

\usepackage{algorithm}
\usepackage{algpseudocode}
\usepackage{subcaption}
\usepackage{newfloat}
\usepackage{adjustbox}
\usepackage{listings}
\usepackage{xcolor}
\usepackage{cleveref}
\usepackage{tikz}
\usepackage{mathtools} % for \coloneqq
\usepackage{amsfonts} % 
\usepackage{pgfplots}
\pgfplotsset{compat=1.18}
\usetikzlibrary[pgfplots.groupplots]

\usetikzlibrary{arrows, automata, positioning}
\tikzset{auto, >=stealth}
\tikzset{every edge/.append style={shorten >= 1pt}}
\tikzset{
    main node/.style={circle,draw,minimum size=1cm,inner sep=0pt},
}

\DeclareCaptionStyle{ruled}{labelfont=normalfont,labelsep=colon,strut=off} % DO NOT CHANGE THIS
\floatstyle{ruled}
\newfloat{listing}{tb}{lst}{}
\floatname{listing}{Listing}

\title{
Specification-Driven Video Search via Foundation Models and Formal Verification
}
\author {
    Yunhao Yang\textsuperscript{\rm 1},
    Jean-Rapha\"el Gaglione\textsuperscript{\rm 1},
    Sandeep Chinchali\textsuperscript{\rm 1},
    Ufuk Topcu\textsuperscript{\rm 1}
}
\affiliations {
    \textsuperscript{\rm 1}The University of Texas at Austin, USA\\
    \{yunhaoyang234, jr.gaglione, sandeepc, utopcu\}@utexas.edu
}

\begin{document}

\maketitle

\begin{abstract}
The increasing abundance of video data enables users to search for events of interest, e.g., emergency incidents. Meanwhile, it raises new concerns, such as the need for preserving privacy. Existing approaches to video search require either manual inspection or a deep learning model with massive training. We develop a method that uses recent advances in vision and language models, as well as formal methods, to search for events of interest in video clips automatically and efficiently. The method consists of an algorithm to map text-based event descriptions into linear temporal logic over finite traces (LTL$_f$) and an algorithm to construct an automaton encoding the video information. Then, the method formally verifies the automaton representing the video against the LTL$_f$ specifications and adds the pertinent video clips to the search result if the automaton satisfies the specifications. We provide qualitative and quantitative analysis to demonstrate the video-searching capability of the proposed method. It achieves over 90 percent precision in searching over privacy-sensitive videos and a state-of-the-art autonomous driving dataset.
\end{abstract}
\glsresetall

% keywords can be removed
% \keywords{First keyword \and Second keyword \and More}

\section{Introduction}

The increasing abundance of video data enables users to search for events of interest, but existing approaches to searching through videos are either inefficient or require manual inspection.
Various cameras, such as security or vehicle dash cameras, gather terabytes of video data, and manually searching for events (e.g., vehicle crashes) over such large-scale data is impractical.
Existing automated video search approaches require either excessive human annotations \cite{video-search-auto} or massive training for neural networks \cite{video-search-challenge, video-search-deep-learning}, which are inefficient.
Furthermore, none of these approaches provide guarantees on the correctness of their search results.

\begin{figure}[t]
    \centering
    \includegraphics[width=\linewidth]{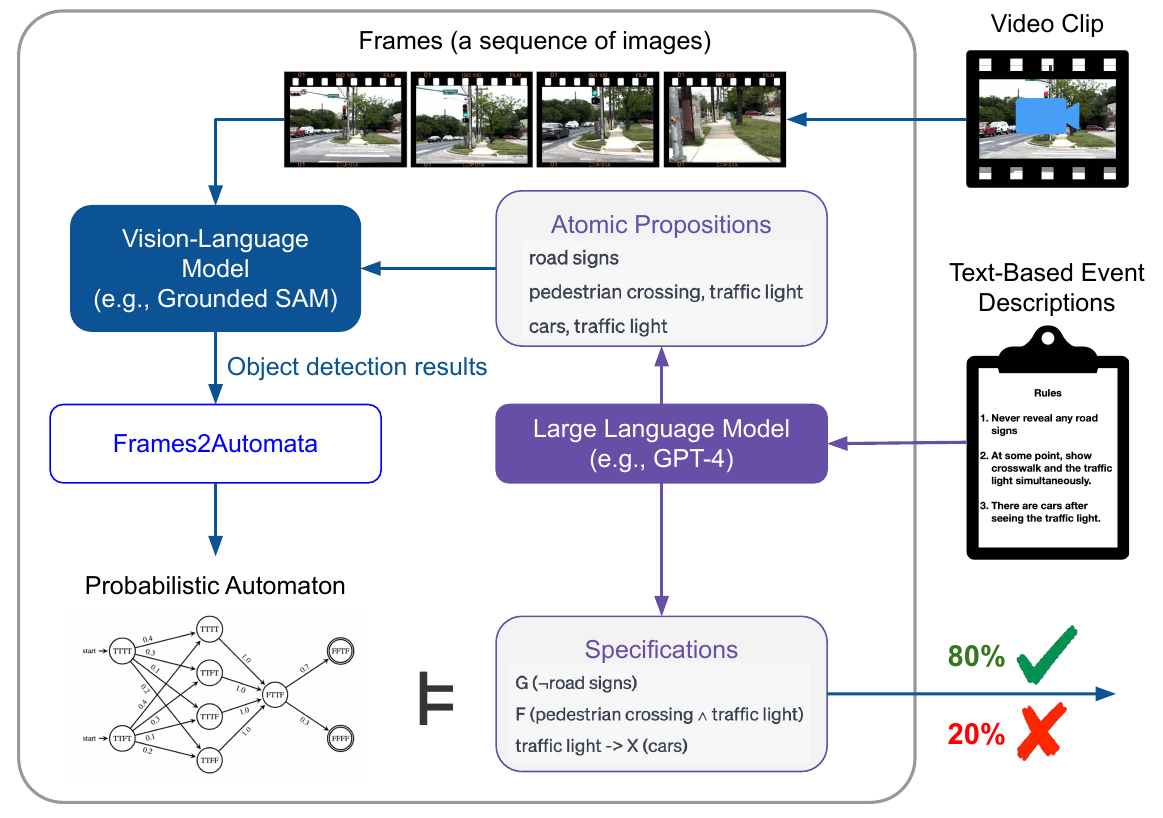}
    \caption{Demonstration of searching an event of interest---described in texts---within a given video.}
    \label{fig: architecture}
\end{figure}

We develop a method that uses recent advances in vision and language models, as well as formal methods, to search for events of interest in video efficiently with probabilistic guarantees. The method includes an algorithm that maps texts to formal specifications and an algorithm to construct automata encoding information from videos, as illustrated in Figure \ref{fig: architecture}.
In practice, many events of interest are expressed in natural language. We accordingly design an algorithm that sends the text-based event description to a text generation model, e.g., GPT-series, and queries the model for the formal specifications and propositions describing the event. This algorithm converts texts to an interpretable format that we can use to search through videos.

Meanwhile, we design an algorithm to construct automata encoding the video information.
The algorithm utilizes a vision-language model to determine whether the event described by the proposition happens in the video, with a score indicating the model's confidence. The algorithm calibrates the confidence to the prediction accuracy of the search query using uncertainty quantification techniques for deep learning models.
It uses the calibrated confidence to compute the probability that the automaton satisfies the specifications.
Hence, we can obtain the probability of each video satisfying the event description during video search and add videos with probabilities above 0.5 to our search result.
In contrast to existing approaches, this method is fully automated, computationally inexpensive, and can provide probabilistic guarantees through formal verification.

We provide case studies and quantitative analysis on privacy-sensitive videos and the state-of-the-art autonomous driving dataset to demonstrate the video-searching capability of the proposed pipeline. We use the ground truth annotations from the datasets to evaluate the proposed pipeline, and the search results achieve over 90 percent precision.

\section{Related Work}

\paragraph{Symbolic Representations.}
Video searching or understanding typically requires the videos to be converted into some symbolic representations encoding the video information.
Existing works \cite{visual-symbolic,long-form-video} utilize object detection models to form symbolic representations of videos that encode object information (e.g., existence, features).
Some other works \cite{JiK0N20,RaiCJDKI0N21,action-space-time} construct graphs that represent each event or object as a temporal sequence of spatial information, which improves the interpretability of the raw data.
However, none of the existing video representations are capable of formal verification. We introduce a video representation that can be formally verified against the provided specifications.

Another work \cite{umili2022groundingltl} uses a deep learning model to classify images and builds a deterministic finite automaton representing an image sequence through the classification results. It then verifies the automaton against temporal logic specifications. This work neither considers potential image classification errors nor provides probabilistic guarantees. In contrast, we incorporate the confidence scores returned by the foundation model to provide probabilistic guarantees. In addition to Umili et al., we propose a method to map natural language to temporal logic specifications.

\paragraph{Video Understanding.}
Video applications like video searching require understanding the content of the videos. Many works focus on short-form video understanding \cite{short1,short2,short3}, i.e., understanding videos less than five seconds, and long-form video understanding \cite{long-form-video,long1,long2,long3}, e.g., movie question-answering.
The existing works interpret the videos to detect actions in the video \cite{deng2023BEAR,BertasiusWT21,liu2021Swin}, track objects \cite{track-obj,track-obj-2}, or detect shot transitions \cite{shot-trans}.
However, these works do not provide any guarantees on their video understanding outcomes. The pipeline we proposed provides guarantees by formally verifying videos against the text-based descriptions of events of interest.

\paragraph{Formal Verification.}
Formal verification is a technique to prove or disprove the satisfaction of a system with respect to certain formal specifications or properties. Such a technique requires a formal representation of the system, such as a finite-state automaton (FSA) or a Markov Decision Process (MDP).
Bai et al. \cite{verify-mdp} propose a method to construct MDPs from driving behaviors and verify the safety of the behaviors. Sai et al. \cite{sai} use MDPs to represent robotic control systems and verify the safety of the system. Yang et al. \cite{Yang2022AutomatonBasedRO} construct FSAs for textual task knowledge extracted from large language models.
None of the existing works apply to videos alone. We take advantage of the emerging foundation models to build formal representations of videos.

\section{Preliminaries}

\paragraph{Probabilistic Automaton.}

We formally define a \gls{aut} \cite{PA} as a tuple
$\Aut =
\langle
    \AutStates,
    \Autstate[init],
    \AutStates[final],
    \AutSymbsOut,
    \AutTransFunc,
    \AutLabelFunc
\rangle$,
where $\AutStates$ is the set of states, $\Autstate[init] \in \AutStates$ is the initial state, $\AutStates[final] \in \AutStates$ is the set of acceptance states, $\AutSymbsOut$ is the set of state labeling symbols, $\AutLabelFunc: \AutStates \to \AutSymbsOut$ is the label function, and $\AutTransFunc: \AutStates \times \AutStates \to \{0, 1\}$
is the transition function indicating whether a given transition is allowed by the automaton.
% The \glspl{aut} transitions are non-deterministic, so the transition function can also be written as a membership function $\AutTransFunc: \AutStates \times \AutStates \to \{0, 1\}$, so that $\AutTransFunc(q, \sigma, q') = 1$ if $q' \in \AutTransFunc(q, \sigma)$. This membership function indicates whether a given transition is allowed by the automaton.
Each transition ($q \xrightarrow{\sigma} q'$) associates with a probability $\mathbb{P}$, which means if we are at state $q$, we choose the transition $q \xrightarrow{\sigma} q'$ with a probability $\mathbb{P}$. Note that for every state, the probabilities of all the outgoing transitions of this state with the input symbol $\sigma$ sum up to 1.

We define a set of atomic propositions $\AutProps$ for the state labeling symbols $\AutSymbsOut \coloneqq 2^{\AutProps}$.
The propositional logic formula is based on these atomic propositions. 
% e.g., $\varphi = \lnot\Autprop[1] \land \Autprop[2]$ where $\Autprop[1],\Autprop[2] \in \AutProps$.

We then define a \emph{trajectory} as a sequence of labels of states and the input symbols of the transitions
\begin{center}
    $\AutLabelFunc(q_0), \AutLabelFunc(q_1), ..., \AutLabelFunc(q_n)$, where $q_i \in Q$. 
\end{center}
The trajectory starts from the initial state $q_0$ and ends at one of the acceptance states $q_n \in \AutStates[final]$. Each trajectory is associated with a probability, which is the product of all the probabilities of the state transitions in the trajectory.

% https://www.ijcai.org/proceedings/2017/0189.pdf
\paragraph{Linear Temporal Logic over Finite Traces.}
Temporal logic represents propositional and first-order logical reasoning with respect to time. 
\Gls{tl} \cite{LTL-f} is a form of temporal logic that deals with finite sequences, i.e., finite-length trajectories. The syntax of \gls{tl} formulas is defined as
\begin{center}
    $
    \varphi \coloneqq \Autprop\in P \mid \lnot\varphi \mid \varphi_1 \land \varphi_2 \mid \lnext\varphi \mid  \varphi_1\luntil\varphi_2
    $.
\end{center}
\Gls{tl} consists of all operations from propositional logic, such as AND ($\land$), OR ($\vee$), XOR ($\oplus$), NOT ($\neg$), IMPLY ($\rightarrow$), etc., and the following temporal operations:
\begin{itemize}
    \item Always ($\lalways \phi_1$): $\phi_1$ is true for every step in the trajectory.
    \item Sometimes ($\leventually \phi_1$): $\phi_1$ is true for at least one step in the trajectory.
    \item Next ($\lnext \phi_1$): $\phi_1$ is true in the next step.
    \item Until ($\phi_1 \luntil \phi_2$): $\phi_1$ has to be true until $\phi_2$ becomes true, $\phi_2$ has to be true at one of the future steps.
\end{itemize}

$\phi_1$ and $\phi_2$ are \gls{tl} formulas. An \gls{tl} formula is composed of variables in $P$ and logic operations, as we call a \emph{specification}. Each formula can be satisfied by a finite sequence of truth valuations of variables in $P$, as we call a trajectory. If a trajectory $T$ satisfies the specification $\Phi$, we denote $T \models \Phi$.

\paragraph{Foundation Model.}
Foundation models are large-scale machine learning models that are trained on a vast amount of data and can be directly applied or fine-tuned to a wide range of downstream tasks.
\Glspl{llm} such as BERT \cite{bert}, CodeX \cite{codex}, and GPT-series \cite{brown2020GPT3,openai2023gpt4} are text foundation models. LLMs are capable of downstream language processing tasks like text generation (next-word prediction), question-answering, text classification and summarization, machine translation, etc. 

Beyond text capability, several multimodal foundation models can understand and process both visual and textual inputs. 
% Such models include multimodal generative models, such as GPT-4 \cite{openai2023gpt4} and Stable Diffusion \cite{stable}, which can generate or modify images according to text inputs. 
CLIP \cite{clip} measures the similarity or consistency between the input texts and images.
Object detection models such as Yolo \cite{yolo}, Grounded-Segment-Anything  (Grounded-SAM) \cite{liu2023grounding,kirillov2023segmentanything}, ViLD \cite{GuLKC22}, GLIP \cite{LiZZYLZWYZHCG22} and R-CNN \cite{rcnn} can detect the existences and positions of objects described in the textual inputs from a given image. 
% Among these object detection models, Yolo and R-CNN are closed-domain, which means they can only recognize a fixed set of objects. Grounded-Segment-Anything, ViLD, and GLIP are open-domain object detection models that can take any text as input and detect the objects described in the text from a given image.
In the later parts, we refer to these models as \glspl{vlm} in the later parts.

\section{Methodology}
We develop a method to search for events of interest in video. For each video clip, the method takes in a set of text-based event descriptions and outputs a score indicating the probability of the description being satisfied in the video. The method has two components: The first is an algorithm to map text-based event descriptions into \gls{tl} specifications. The second component is an algorithm to construct a probabilistic automaton encoding the video information.
Then, the method verifies the automaton against the specifications to obtain a probability that the automaton satisfies the \gls{tl} specifications.
By applying this method to a set of videos, we can efficiently search for events of interest in these videos by finding all the videos whose automata satisfy the specifications.

\subsection{Text-Based Description to \gls{tl} Specification}
We develop an algorithm to map text-based descriptions of events of interest to \gls{tl} specifications.
The algorithm first extracts a set of atomic propositions and by querying the \gls{llm} to extract \emph{noun phrases} from the texts:
\vspace{0.2cm}
\begin{lstlisting}[language=completion]
    <prompt>Extract noun phrases from the following rules: 
    <emph>A list of rules.</emph></prompt><completion> 
    1. <emph>noun phrase 1</emph>
    2. <emph>noun phrase 2</emph>...</completion>
\end{lstlisting}
\vspace{0.2cm}
Note that a noun phrase is a group of words that functions as a single unit within a sentence and centers around a noun.
We consider these noun phrases as atomic propositions. Then, the algorithm queries the \gls{llm} again to transform the textual rules to a set of \gls{tl} specifications with respect to the atomic propositions:
\vspace{0.2cm}
\begin{lstlisting}[language=completion]
    <prompt>Define the following rules in temporal logic with atomic propositions <emph>noun phrase 1</emph>, <emph>noun phrase 2</emph>, ...: 
    <emph>A list of rules.</emph></prompt><completion> 
    1. <emph>Temporal logic formula 1</emph>
    2. <emph>Temporal logic formula 2</emph>...</completion>
\end{lstlisting}
\vspace{0.2cm}
Now we have the sequence of frames $\mathcal{F}$, a set of atomic propositions $P$, and \gls{tl} specifications $\Phi$.

\subsection{Video to Probabilistic Automaton}
We develop an algorithm to extract information from videos and construct probabilistic automata encoding those information.
The algorithm starts from a provided video and a set of textual rules regarding the video. We extract frames from the video at regular intervals, where each frame is an image. We denote the number of frames extracted in each second of the video as \emph{frame frequency}, with a unit of \emph{frames per second}.

\paragraph{Calibrating Confidence and Accuracy}
We use a \gls{vlm}, e.g., an open-domain object detection model, to evaluate each atomic proposition at each frame. In particular, a \gls{vlm} that can take text-based propositions and the frame as inputs and returns \emph{confidence score} for each proposition. 
A confidence score is a softmax score that indicates how certain the model is about its prediction. These scores are between 0 and 1, where a value close to 1 indicates high confidence and a value close to 0 means low confidence.

The model detects whether the object or scene described by the atomic proposition appears in the image with a certain confidence. We evaluate the proposition as true only if the object or scene is detected.
However, the \gls{vlm} can raise detection errors. We consider a detection result to be \emph{correct} only if the \gls{vlm} detected an object that actually exists in the image.
Therefore, we utilize the confidence scores returned by the model and estimate a mapping between confidence scores and the percentage of correct detection over all the detection results, as we call \emph{classification accuracy}.
\begin{definition}
    Let $N$ be the number of detection results whose confidence scores returned by the \gls{vlm} fall into a particular range $[c_1, c_2)$, $C$ be the number of correct detection results (the detected object actually appears in the image) with confidence scores between $[c_1, c_2)$, we define the \textsc{classification accuracy} $A_C$ of confidence interval $(c_1, c_2]$ as
    \begin{center}
        $A_C = \frac{C}{N}$.
    \end{center}
\end{definition}

We assume that a \gls{vlm} performs consistently on data outside its training dataset.
Under this assumption, we consider the classification accuracy at each confidence interval on a validation dataset as the probability of the detection result being correct on realistic data.
Therefore, we apply the model to an image classification task on the validation dataset to estimate a confidence-accuracy mapping and assume it also applies to realistic data.

We apply the \gls{vlm} on a validation dataset independent of the video to obtain the classification accuracies at each confidence interval.
Hence, we get a set of confidence-accuracy pairs.
Then, we use a logistic function 
\begin{equation}
    f(x) = \frac{1}{1 + \exp(-k \cdot (x - x_0))}
\end{equation}
to estimate a mapping function $\mathcal{M: C \mapsto A}$ that maps confidence scores to an accuracies. We use these accuracies in constructing probabilistic automata.

\paragraph{Constructing Probabilistic Automaton}
We now have a \gls{vlm} $M_V: \mathcal{F} \times P \mapsto C$ that takes in a frame and a proposition and returns a confidence score, a sequence of frames $\mathcal{F}$, atomic propositions $P$, and a mapping function $\mathcal{M}$. We use them to construct a probabilistic automaton and verify it against the \gls{tl} specifications $\Phi$.

We start by creating an initial state $q_0$ whose label is \emph{none}, meaning that the initial state's label will not be counted into the trajectory during verification. Then, we process the sequence of frames in order to construct a probabilistic automaton. This procedure includes four steps:

We process each frame in four steps: We first send the frame and all atomic propositions to the \gls{vlm} and obtain a list of confidence scores associated with the propositions. In the second step, we map the confidence scores into accuracies through the function $\mathcal{M}$. In the third step, we create $2^{|P|}$ ($|P|$ is the number of atomic propositions) new automaton states, each corresponding to a conjunction of propositions in $2^P$. The corresponding conjunction of propositions is the label for the state. As an example, if an atomic proposition set $P = \{p_1, p_2\}$, we build four states with labels $p_1 \land p_2$, $\neg p_1 \land p_2$, $p_1 \land \neg p_2$, and $\neg p_1 \land \neg B$, respectively.

In the fourth step, we compute a probability score for each newly created state at frame $\mathcal{F}_j$. For a new state $q_{j,k}$, the probability is the product of the probabilities for all the propositions
\begin{equation}
    \mathbb{P}_{j,k} = \prod_{p_i} \mathcal{M} ( M_V(\mathcal{F}_j, p_i) ) \quad \text{for all } p_i \in P,
\end{equation}
where $\mathcal{F}_j$ is the current frame. The probability of the negation of a proposition ($\neg p_i$) is $1 - \mathcal{M} ( M_V(\mathcal{F}_j, p_i) )$. For every new state $q_{j,k}$ whose probability $\mathbb{P}_{j,k} > 0$, we construct transitions from all the states from the previous frame $\mathcal{F}_{j-1}$ to $q_{j,k}$ with probability $\mathbb{P}_{j,k}$. After constructing the transitions, we remove all the new states that do not have incoming transitions. If $\mathcal{F}_j$ is the first frame (j=1), we add transitions from the initial state $q_0$ to every new state.

We repeat the four steps for all the frames in $\mathcal{F}$ and add the states created from the last frame to the set of acceptance states. Hence, we complete the automaton construction. We present the complete procedure in Algorithm \ref{alg: frames2automata}.

However, the resulting automaton will consist of $|\mathcal{F}| \times 2^{|P|}$ ($|\mathcal{F}|$ is the number of frames) states, which could lead to high computational complexity. We set two thresholds $t_T$ and $t_F$ and modify the mapping function to
\begin{equation}
    \mathcal{M}'(c) = 
    \begin{cases}
        1 \text{ if } c \ge t_T \\
        \mathcal{M}(c) \text{ if } t_F < c < t_T \\
        0 \text{ if } c \le t_F,
    \end{cases}
\end{equation}
where $c \in C$ is the confidence score. We use $\mathcal{M}'$ as the mapping function for Algorithm \ref{alg: frames2automata} instead of $\mathcal{M}$. By doing so, we can eliminate a proportion of states due to zero-probability incoming transitions.

\begin{algorithm}[t]
  \caption{Automaton Construction from Video Frames}\label{alg: frames2automata}
  \begin{algorithmic}[1]
    \Procedure{\textbf{Frames2Automata}}{Vision-language model $M_V$, Frames $\mathcal{F}$, Output propositions $P$, Mapping function $\mathcal{M}$, True threshold $t_T$, False threshold $t_F$}

    \State $\AutSymbsIn, \AutSymbsOut, \AutStates, \AutStates[final], \AutLabelFunc, \AutTransFunc$ = $A$, $2^P$, [$\Autstate[init]$], [], [$(\Autstate[init]: none)$], []
    \State prev = [$\Autstate[init]$]

    \For{$\mathcal{F}_j$ in $\mathcal{F}$}
        \State probability = dictionary()
        \For{$p_i$ in $P$}
            \State $c = M_V(\mathcal{F}_j, p_i)$
            \State probability[$p_i$] = $\mathcal{M}(c)$
            \State probability[$\neg p_i$] = $1 - \mathcal{M}(c)$
        \EndFor
        
        \State current = []
        \For{$conj_k$ in conjunctions of $2^P$}
            \State $\mathbb{P}_{j,k} = \prod_{p_i \in conj_k}$ probability[$p_i$]
            \If{$\mathbb{P}_{j,k} > 0$}
                \State $\AutStates$.append($q_{j,k}$)
                \State current.append($q_{j,k}$)
                \State $\AutLabelFunc$.append(($q_{j,k}: conj_k$))
                \State $\AutStates[final]$.append($q_{j,k}$) if $j = |\mathcal{F}|$
                \For{$q_{j-1}$ in current}
                    \State $\AutTransFunc$.append(($q_{j-1}, q_{j,k}, \mathbb{P}_{j,k}$))
                \EndFor
            \EndIf
        \EndFor
        
        \State prev = current
    \EndFor

    \State \textbf{return} $\AutStates, \AutStates[init], \AutStates[final], \AutSymbsIn, \AutSymbsOut, \AutTransFunc, \AutLabelFunc$ 
    \EndProcedure
  \end{algorithmic}
\end{algorithm}

\subsection{Verification and Video Search}
After we construct probabilistic automata for all the videos and convert event descriptions into \gls{tl} specifications, we verify each automaton against the specification.
If an automaton satisfies all the specifications with probability above a certain threshold (typically 0.5), we add the video corresponding to this automaton to our search result.
By doing so, we can efficiently obtain all the videos containing the provided events of interest with probabilistic guarantees.

\section{Empirical Demonstration}
We empirically demonstrate the video search method on multiple proof-of-concept examples. For each video clip, we verify it against the provided specifications and add it to the search result if the verification probability is above 50 percent. Then, we provide quantitative analysis on a privacy-annotated video dataset and on a state-of-the-art autonomous driving dataset. Both datasets consist of ground truth annotations which we can use to evaluate the performance of our video search results.

\subsection{Confidence-Accuracy Mapping: Grounded-SAM}
We choose an open-domain object detection model---Grounded-SAM \cite{kirillov2023segmentanything,liu2023grounding}---as the \gls{vlm} for the experiments and estimate the confidence-probability mapping on a validation dataset ImageNet \cite{imagenet}.

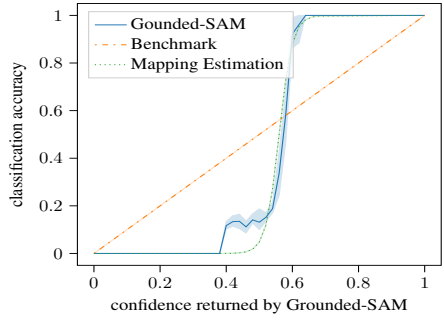
\begin{figure}[t]
    \centering
    \resizebox{0.7\linewidth}{0.5\linewidth}{% This file was created with tikzplotlib v0.10.1.
\begin{tikzpicture}

\definecolor{darkgray176}{RGB}{176,176,176}
\definecolor{darkorange25512714}{RGB}{255,127,14}
\definecolor{forestgreen4416044}{RGB}{44,160,44}
\definecolor{steelblue31119180}{RGB}{31,119,180}
\definecolor{lightgray204}{RGB}{204,204,204}

\begin{axis}[
legend cell align={left},
legend style={
  fill opacity=0.8,
  draw opacity=1,
  text opacity=1,
  at={(0.03,0.97)},
  anchor=north west,
  draw=lightgray204
},
tick align=outside,
tick pos=left,
% title=Grounded-SAM as Image Classifier,
x grid style={darkgray176},
xlabel={confidence returned by Grounded-SAM},
xmin=-0.05, xmax=1.05,
xtick style={color=black},
y grid style={darkgray176},
ylabel={classification accuracy},
ymin=-0.05, ymax=1.05,
ytick style={color=black}
]
\path [draw=steelblue31119180, fill=steelblue31119180, opacity=0.2]
(axis cs:0,0)
--(axis cs:0,0)
--(axis cs:0.3,0)
--(axis cs:0.32,0)
--(axis cs:0.34,0)
--(axis cs:0.36,0)
--(axis cs:0.38,0)
--(axis cs:0.4,0.101932982061453)
--(axis cs:0.42,0.114087770204917)
--(axis cs:0.44,0.103942474584676)
--(axis cs:0.46,0.0860344827586207)
--(axis cs:0.48,0.126917148362235)
--(axis cs:0.5,0.100448394587237)
--(axis cs:0.52,0.134320881637326)
--(axis cs:0.54,0.170289855072464)
--(axis cs:0.56,0.24375)
--(axis cs:0.58,0.468362282878412)
--(axis cs:0.6,0.863354037267081)
--(axis cs:0.62,0.884615384615385)
--(axis cs:0.64,1)
--(axis cs:0.66,1)
--(axis cs:0.68,1)
--(axis cs:1,1)
--(axis cs:1,1)
--(axis cs:1,1)
--(axis cs:0.68,1)
--(axis cs:0.66,1)
--(axis cs:0.64,1)
--(axis cs:0.62,1)
--(axis cs:0.6,0.990740740740741)
--(axis cs:0.58,0.677777777777778)
--(axis cs:0.56,0.427335016227912)
--(axis cs:0.54,0.214673913043478)
--(axis cs:0.52,0.171270247229327)
--(axis cs:0.5,0.187352245862884)
--(axis cs:0.48,0.167130253379113)
--(axis cs:0.46,0.135513637105994)
--(axis cs:0.44,0.164754086004086)
--(axis cs:0.42,0.158404382104877)
--(axis cs:0.4,0.134681460272011)
--(axis cs:0.38,0)
--(axis cs:0.36,0)
--(axis cs:0.34,0)
--(axis cs:0.32,0)
--(axis cs:0.3,0)
--(axis cs:0,0)
--cycle;

\path [draw=darkorange25512714, fill=darkorange25512714, opacity=0.2]
(axis cs:0,0)
--(axis cs:0,0)
--(axis cs:0.3,0.3)
--(axis cs:0.32,0.32)
--(axis cs:0.34,0.34)
--(axis cs:0.36,0.36)
--(axis cs:0.38,0.38)
--(axis cs:0.4,0.4)
--(axis cs:0.42,0.42)
--(axis cs:0.44,0.44)
--(axis cs:0.46,0.46)
--(axis cs:0.48,0.48)
--(axis cs:0.5,0.5)
--(axis cs:0.52,0.52)
--(axis cs:0.54,0.54)
--(axis cs:0.56,0.56)
--(axis cs:0.58,0.58)
--(axis cs:0.6,0.6)
--(axis cs:0.62,0.62)
--(axis cs:0.64,0.64)
--(axis cs:0.66,0.66)
--(axis cs:0.68,0.68)
--(axis cs:1,1)
--(axis cs:1,1)
--(axis cs:1,1)
--(axis cs:0.68,0.68)
--(axis cs:0.66,0.66)
--(axis cs:0.64,0.64)
--(axis cs:0.62,0.62)
--(axis cs:0.6,0.6)
--(axis cs:0.58,0.58)
--(axis cs:0.56,0.56)
--(axis cs:0.54,0.54)
--(axis cs:0.52,0.52)
--(axis cs:0.5,0.5)
--(axis cs:0.48,0.48)
--(axis cs:0.46,0.46)
--(axis cs:0.44,0.44)
--(axis cs:0.42,0.42)
--(axis cs:0.4,0.4)
--(axis cs:0.38,0.38)
--(axis cs:0.36,0.36)
--(axis cs:0.34,0.34)
--(axis cs:0.32,0.32)
--(axis cs:0.3,0.3)
--(axis cs:0,0)
--cycle;

\path [draw=forestgreen4416044, fill=forestgreen4416044, opacity=0.2]
(axis cs:0,6.9144001069354e-13)
--(axis cs:0,6.9144001069354e-13)
--(axis cs:0.3,2.26032429790357e-06)
--(axis cs:0.32,6.14417460221471e-06)
--(axis cs:0.34,1.67014218480951e-05)
--(axis cs:0.36,4.53978687024342e-05)
--(axis cs:0.38,0.000123394575986232)
--(axis cs:0.4,0.000335350130466478)
--(axis cs:0.42,0.000911051194400642)
--(axis cs:0.44,0.00247262315663477)
--(axis cs:0.46,0.00669285092428483)
--(axis cs:0.48,0.0179862099620915)
--(axis cs:0.5,0.0474258731775667)
--(axis cs:0.52,0.119202922022117)
--(axis cs:0.54,0.268941421369995)
--(axis cs:0.56,0.5)
--(axis cs:0.58,0.731058578630005)
--(axis cs:0.6,0.880797077977882)
--(axis cs:0.62,0.952574126822433)
--(axis cs:0.64,0.982013790037908)
--(axis cs:0.66,0.993307149075715)
--(axis cs:0.68,0.997527376843365)
--(axis cs:1,0.999999999721053)
--(axis cs:1,0.999999999721053)
--(axis cs:1,0.999999999721053)
--(axis cs:0.68,0.997527376843365)
--(axis cs:0.66,0.993307149075715)
--(axis cs:0.64,0.982013790037908)
--(axis cs:0.62,0.952574126822433)
--(axis cs:0.6,0.880797077977882)
--(axis cs:0.58,0.731058578630005)
--(axis cs:0.56,0.5)
--(axis cs:0.54,0.268941421369995)
--(axis cs:0.52,0.119202922022117)
--(axis cs:0.5,0.0474258731775667)
--(axis cs:0.48,0.0179862099620915)
--(axis cs:0.46,0.00669285092428483)
--(axis cs:0.44,0.00247262315663477)
--(axis cs:0.42,0.000911051194400642)
--(axis cs:0.4,0.000335350130466478)
--(axis cs:0.38,0.000123394575986232)
--(axis cs:0.36,4.53978687024342e-05)
--(axis cs:0.34,1.67014218480951e-05)
--(axis cs:0.32,6.14417460221471e-06)
--(axis cs:0.3,2.26032429790357e-06)
--(axis cs:0,6.9144001069354e-13)
--cycle;

\addplot [semithick, steelblue31119180]
table {%
0 0
0.3 0
0.32 0
0.34 0
0.36 0
0.38 0
0.4 0.116910280066119
0.42 0.133765098890504
0.44 0.134855241735976
0.46 0.111633045554636
0.48 0.140978763543412
0.5 0.131193464433098
0.52 0.151819706930288
0.54 0.188128512274475
0.56 0.337042463036423
0.58 0.584610789449499
0.6 0.922417759374281
0.62 0.961538461538462
0.64 1
0.66 1
0.68 1
1 1
};
\addlegendentry{Gounded-SAM}
\addplot [semithick, darkorange25512714, dash pattern=on 1pt off 3pt on 3pt off 3pt]
table {%
0 0
0.3 0.3
0.32 0.32
0.34 0.34
0.36 0.36
0.38 0.38
0.4 0.4
0.42 0.42
0.44 0.44
0.46 0.46
0.48 0.48
0.5 0.5
0.52 0.52
0.54 0.54
0.56 0.56
0.58 0.58
0.6 0.6
0.62 0.62
0.64 0.64
0.66 0.66
0.68 0.68
1 1
};
\addlegendentry{Benchmark}
\addplot [semithick, forestgreen4416044, dotted]
table {%
0 6.9144001069354e-13
0.3 2.26032429790357e-06
0.32 6.14417460221471e-06
0.34 1.67014218480951e-05
0.36 4.53978687024342e-05
0.38 0.000123394575986232
0.4 0.000335350130466478
0.42 0.000911051194400642
0.44 0.00247262315663477
0.46 0.00669285092428483
0.48 0.0179862099620915
0.5 0.0474258731775667
0.52 0.119202922022117
0.54 0.268941421369995
0.56 0.5
0.58 0.731058578630005
0.6 0.880797077977882
0.62 0.952574126822433
0.64 0.982013790037908
0.66 0.993307149075715
0.68 0.997527376843365
1 0.999999999721053
};
\addlegendentry{Mapping Estimation}
\end{axis}

\end{tikzpicture}}
    \caption{Confidence score returned by the Grounded-SAM versus its classification accuracy (blue line). The estimated mapping function (green dotted line) is $\mathcal{M}(x) = 1/(1 + \exp(-50 \cdot (x - 0.56)))$. The orange line is the benchmark mapping, which is an identity function $\overline{\mathcal{M}} (x) = x$.}
    \label{fig: conf-acc}
\end{figure}

We send each image and the complete list of labels from the validation dataset to the Grounded-SAM to obtain a confidence score with a label for each image. If the object detected by the Grounded-SAM is identical to the image's label, then we consider this case as a correct prediction.
Figure \ref{fig: conf-acc} shows the confidence-accuracy mapping. We estimate this mapping using a logistic function
\begin{equation}
    \label{eq: mapping}
    \mathcal{M}(x) = \frac{1}{1 + \exp(-50 \cdot (x - 0.56))},
\end{equation}
which will be used for all the experiments in the later sections.

According to Figure \ref{fig: conf-acc}, the classification accuracy is consistently equal to one when the confidence is greater than 0.64 and consistently equal to zero when the confidence is less than 0.38. For the purpose of simplifying the constructed automaton, we set a true threshold $t_T = 0.64$ and a false threshold $t_F = 0.38$.

\subsection{Proof-of-Concept Demonstrations}
We first demonstrate the proposed method through two proof-of-concept examples, where we search through recorded videos that satisfy manually generated rules. 
\begin{figure}[t]
    \centering
    \includegraphics[width=0.8\linewidth]{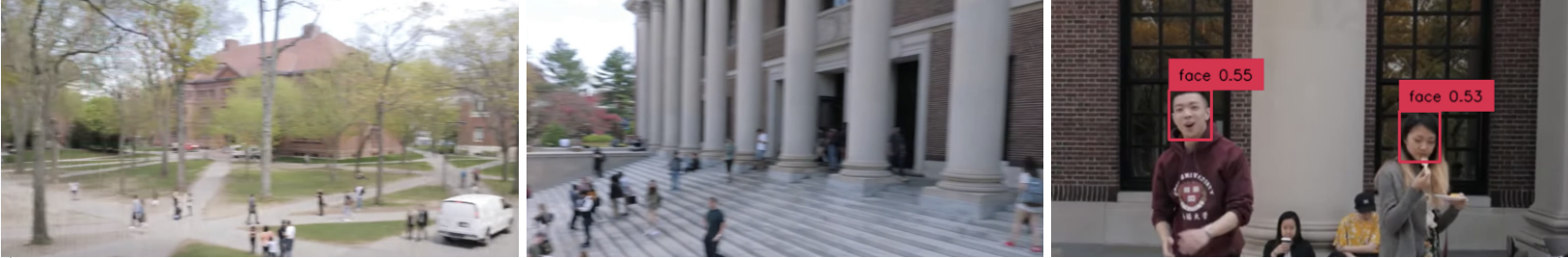}
    \includegraphics[width=0.8\linewidth]{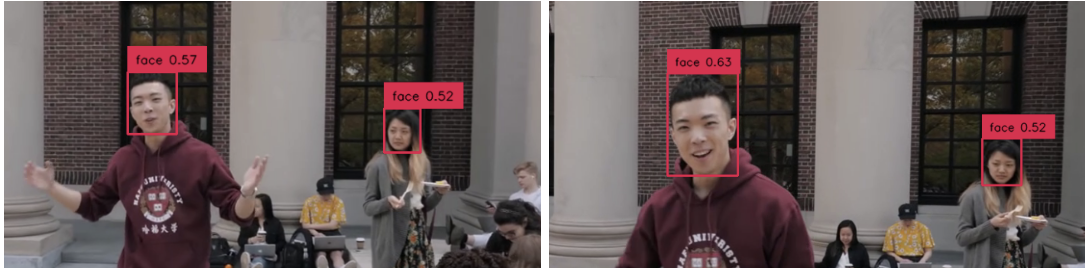}
    \caption{Object detection results on the frames from college introductory videos.}
    \label{fig: school}
\end{figure}

\begin{figure}[t]
    \centering
    \begin{tikzpicture}[thick,scale=.6, node distance=2.2cm, every node/.style={transform shape}]
	\node[state, initial] (0) at (0, 0) {-};
	\node[state] (1) at (2, 0) {F};
    \node[state] (2) at (4, 0) {F};
    \node[state] (31) at (6, 1) {T};
	\node[state] (32) at (6, -1) {F};
	\node[state] (41) at (9, 1) {T};
    \node[state] (42) at (9, -1) {F};
    \node[state, accepting] (51) at (12, 1) {T};
    \node[state, accepting] (52) at (12, -1) {F};

	\draw[->, shorten >=1pt, sloped] (0) to[left] node[above, align=center] {1.0} (1);
    \draw[->, shorten >=1pt, sloped] (1) to[left] node[above, align=center] {1.0} (2);
    \draw[->, shorten >=1pt, sloped] (2) to[left] node[above, align=center] {0.38} (31);
    \draw[->, shorten >=1pt, sloped] (2) to[left] node[above, align=center] {0.62} (32);

    \draw[->, shorten >=1pt, sloped] (31) to[left] node[above, align=center] {0.62} (41);
    \draw[->, shorten >=1pt, sloped] (32) to[left] node[below, align=center, xshift=-8mm] {0.62} (41);
    \draw[->, shorten >=1pt, sloped] (31) to[left] node[below, align=center, xshift=-8mm] {0.38} (42);
    \draw[->, shorten >=1pt, sloped] (32) to[left] node[below, align=center] {0.38} (42);

    \draw[->, shorten >=1pt, sloped] (41) to[left] node[above, align=center] {0.97} (51);
    \draw[->, shorten >=1pt, sloped] (42) to[left] node[below, align=center, xshift=-8mm] {0.97} (51);
    \draw[->, shorten >=1pt, sloped] (41) to[left] node[below, align=center, xshift=-8mm] {0.03} (52);
    \draw[->, shorten >=1pt, sloped] (42) to[left] node[below, align=center] {0.03} (52);

 %    \node[state, initial] (s0) at (0, -3) {-};
	% \node[state] (s1) at (2, -3) {F};
 %    \node[state] (s2) at (4, -3) {F};
 %    \node[state] (s3) at (6, -3) {F};
	% \node[state] (s4) at (9, -3) {F};
 %    \node[state, accepting] (s5) at (12, -3) {F};

 %    \draw[->, shorten >=1pt, sloped] (s0) to[left] node[above, align=center] {1.0} (s1);
 %    \draw[->, shorten >=1pt, sloped] (s1) to[left] node[above, align=center] {1.0} (s2);
 %    \draw[->, shorten >=1pt, sloped] (s2) to[left] node[above, align=center] {1.0} (s3);
 %    \draw[->, shorten >=1pt, sloped] (s3) to[left] node[above, align=center] {1.0} (s4);
 %    \draw[->, shorten >=1pt, sloped] (s4) to[left] node[above, align=center] {1.0} (s5);

 %    \node[state, initial] (m0) at (0, -4.5) {-};
	% \node[state] (m1) at (2, -4.5) {T};
 %    \node[state] (m2) at (4, -4.5) {F};
 %    \node[state] (m3) at (6, -4.5) {T};
	% \node[state] (m4) at (9, -4.5) {F};
 %    \node[state, accepting] (m5) at (12, -4.5) {T};

 %    \draw[->, shorten >=1pt, sloped] (m0) to[left] node[above, align=center] {1.0} (m1);
 %    \draw[->, shorten >=1pt, sloped] (m1) to[left] node[above, align=center] {1.0} (m2);
 %    \draw[->, shorten >=1pt, sloped] (m2) to[left] node[above, align=center] {1.0} (m3);
 %    \draw[->, shorten >=1pt, sloped] (m3) to[left] node[above, align=center] {1.0} (m4);
 %    \draw[->, shorten >=1pt, sloped] (m4) to[left] node[above, align=center] {1.0} (m5);
 
\end{tikzpicture}
    \caption{Automaton corresponding to the frames of the video in Figure \ref{fig: school}. `T' and `F' in each state indicate the state's label, either ``faces = True" or ``faces = False."}
    \label{fig: face}
\end{figure}
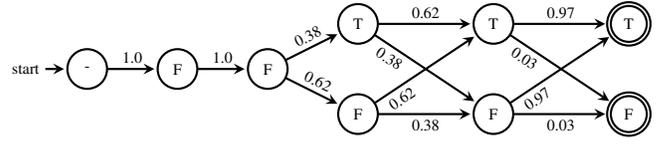

\paragraph{Single-Rule Verification on College Introductory Videos}
We start with an example that searches for videos that satisfy a single privacy rule: never show faces in the video.
We follow the procedure of processing textual rules to \gls{tl} specifications and obtain the specification $Phi$, where
\begin{center}
    $\Phi = \lalways \neg faces$.
\end{center}
During this procedure, we query GPT-4 \cite{openai2023gpt4} to transform textual rules into \gls{tl} specifications.

We randomly select college introductory videos from YouTube and break each video into frames with a frequency of one frame per second. Next, we apply the Grounded-SAM object detection model to detect ``faces" in each frame and present the detection results in Figure \ref{fig: school}.
Then, we follow Algorithm \ref{alg: frames2automata} to construct an automaton from the frames of each video. 

Figure \ref{fig: face} shows one sample automaton constructed from the frames of a college introductory video in Figure \ref{fig: school}.
We use a probabilistic model checker implemented by Stormpy \cite{stormpy} to compute the probability that $\Phi$ is satisfied.
The model checker returns a probability of $0.7\%$, which is the probability of the video satisfying $\Phi$. This means the video in Figure \ref{fig: school} will not be added to the search result since its probability of satisfying $\Phi$ is below 0.5. We repeat the verification procedure to all other college introductory videos and add videos whose probability of satisfying $\Phi$ is greater than 0.5 to the search result. We present more college introductory videos in Appendix \ref{sec: college}.

\paragraph{Multiple-Rule Verification on Traffic Recordings}
\begin{figure}[t]
    \centering
    \includegraphics[width=\linewidth]{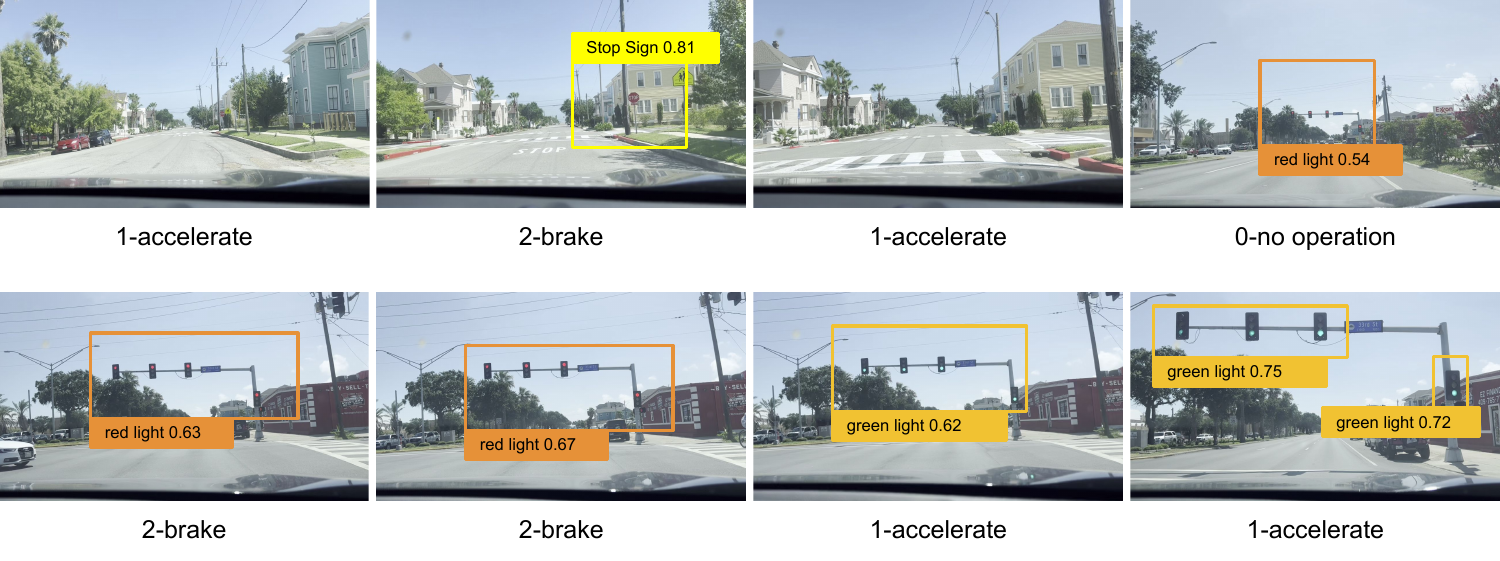}
    \caption{Sample frames from the Driving Control Dataset with the object detection results. We present the label under each frame.}
    \label{fig: driving}
\end{figure}

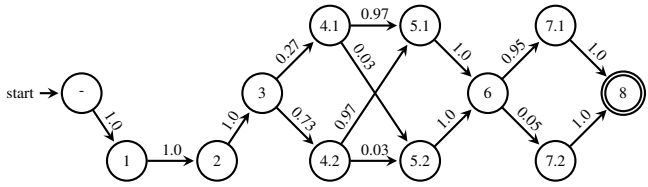
\begin{figure}[t]
    \centering
    \begin{tikzpicture}[thick,scale=.6, node distance=2.2cm, every node/.style={transform shape}]
	\node[state, initial] (0) at (-2, 0) {-};
	\node[state] (1) at (-1, -1.5) {1};
    \node[state] (2) at (1, -1.5) {2};
    \node[state] (3) at (2, 0) {3};
	\node[state] (41) at (3.5, 1.5) {4.1};
    \node[state] (42) at (3.5, -1.5) {4.2};
    \node[state] (51) at (5.5, 1.5) {5.1};
    \node[state] (52) at (5.5, -1.5) {5.2};
    \node[state] (6) at (7, 0) {6};
    \node[state] (71) at (8.5, 1.5) {7.1};
    \node[state] (72) at (8.5, -1.5) {7.2};
    \node[state, accepting] (8) at (10, 0) {8};

	\draw[->, shorten >=1pt, sloped] (0) to[left] node[above, align=center] {1.0} (1);
    \draw[->, shorten >=1pt, sloped] (1) to[left] node[above, align=center] {1.0} (2);
    \draw[->, shorten >=1pt, sloped] (2) to[left] node[above, align=center] {1.0} (3);

    \draw[->, shorten >=1pt, sloped] (3) to[left] node[above, align=center] {0.27} (41);
    \draw[->, shorten >=1pt, sloped] (3) to[left] node[above, align=center] {0.73} (42);

    \draw[->, shorten >=1pt, sloped] (41) to[left] node[above, align=center] {0.97} (51);
    \draw[->, shorten >=1pt, sloped] (42) to[left] node[above, align=center, xshift=-8mm] {0.97} (51);
    \draw[->, shorten >=1pt, sloped] (41) to[left] node[above, align=center, xshift=-8mm] {0.03} (52);
    \draw[->, shorten >=1pt, sloped] (42) to[left] node[above, align=center] {0.03} (52);

    \draw[->, shorten >=1pt, sloped] (51) to[left] node[above, align=center] {1.0} (6);
    \draw[->, shorten >=1pt, sloped] (52) to[left] node[above, align=center] {1.0} (6);

    \draw[->, shorten >=1pt, sloped] (6) to[left] node[above, align=center] {0.95} (71);
    \draw[->, shorten >=1pt, sloped] (6) to[left] node[above, align=center] {0.05} (72);

    \draw[->, shorten >=1pt, sloped] (71) to[left] node[above, align=center] {1.0} (8);
    \draw[->, shorten >=1pt, sloped] (72) to[left] node[above, align=center] {1.0} (8);
\end{tikzpicture}
    \caption{Automaton constructed from the frames in Figure \ref{fig: driving}. The labels for each state are in Table \ref{tab: label}.
    }
    \label{fig: drive-aut}
\end{figure}

\begin{table}[t]
    \centering
    \begin{tabular}{||c|c|c|c||}
    \hline
    State & Label & State & Label\\
    \hline
    1,3,7.2 & accelerate & 2 & brake, stop sign \\
    4.2 & no-op & 4.1 & no-op, red light \\
    5.1,6 & brake, red light & 5.2 & brake \\
    7.1, 8 & accelerate, green & & \\
    \hline
    \end{tabular}
    \caption{State labels of the automaton in Figure \ref{fig: drive-aut}.}
    \label{tab: label}
\end{table}
We have so far demonstrated the algorithm's capability in single-rule verification. We now search for traffic recording clips that satisfy multiple rules. We send the following prompt to GPT-4 to obtain the \gls{tl} specifications regarding the traffic rules:
\vspace{0.2cm}
\begin{lstlisting}[language=completion]
    Given the propositions: pedestrian, stop sign, red light, green light, accelerate, and brake.
    Transform the following rules into temporal logic formulas:
    1. Yield to pedestrians.
    2. Slow down after seeing the stop sign or red light.
    3. Accelerate sometime after the light turns from red to green.
\end{lstlisting}
\vspace{0.2cm}
The specifications are

$\Phi_1 = \lalways$ (pedestrian $\rightarrow$ brake),

$\Phi_2 = \lalways$ ((stop sign $\vee$ red light) $\rightarrow \lnext$ brake),

$\Phi_3 = \lalways$ ((red light $\land \lnext$ green light) $\rightarrow \lnext ( \leventually$ accelerate) ).

We collect a set of driving recordings and extract one frame per second from the recordings. Then, we manually label each frame with driving operations 1-accelerate, 2-brake, or 0-no operation and construct a small-scale \emph{Driving Control Dataset}. We present some sample data from the dataset in Appendix \ref{sec: dataset}.

We apply the method to find recording clips that satisfy our selected traffic rules. We have the set of symbols $\AutSymbsOut = \{$ accelerate, brake, no operation, pedestrian, stop sign, red light, green light $\}$, where the first three symbols are labels from the dataset, and the others are evaluated through the \gls{vlm}. An example of the frames from the dataset is presented in Figure \ref{fig: driving}, and the corresponding constructed automaton is presented in Figure \ref{fig: drive-aut}. 

In this example, the probabilities of the three specifications being satisfied are 100\%, 73\%, and 100\%, respectively. Since this recording clip satisfies all three specifications with probabilities greater than 0.5, we add this recording clip to the search result. We can apply this procedure to search for video clips from a long video, as presented in Appendix \ref{sec: dataset}.

\subsection{Quantitative Analysis over Realistic Datasets}
\paragraph{Search over Privacy-Sensitive Videos.}
\begin{figure}[t]
    \centering
    \includegraphics[width=\linewidth]{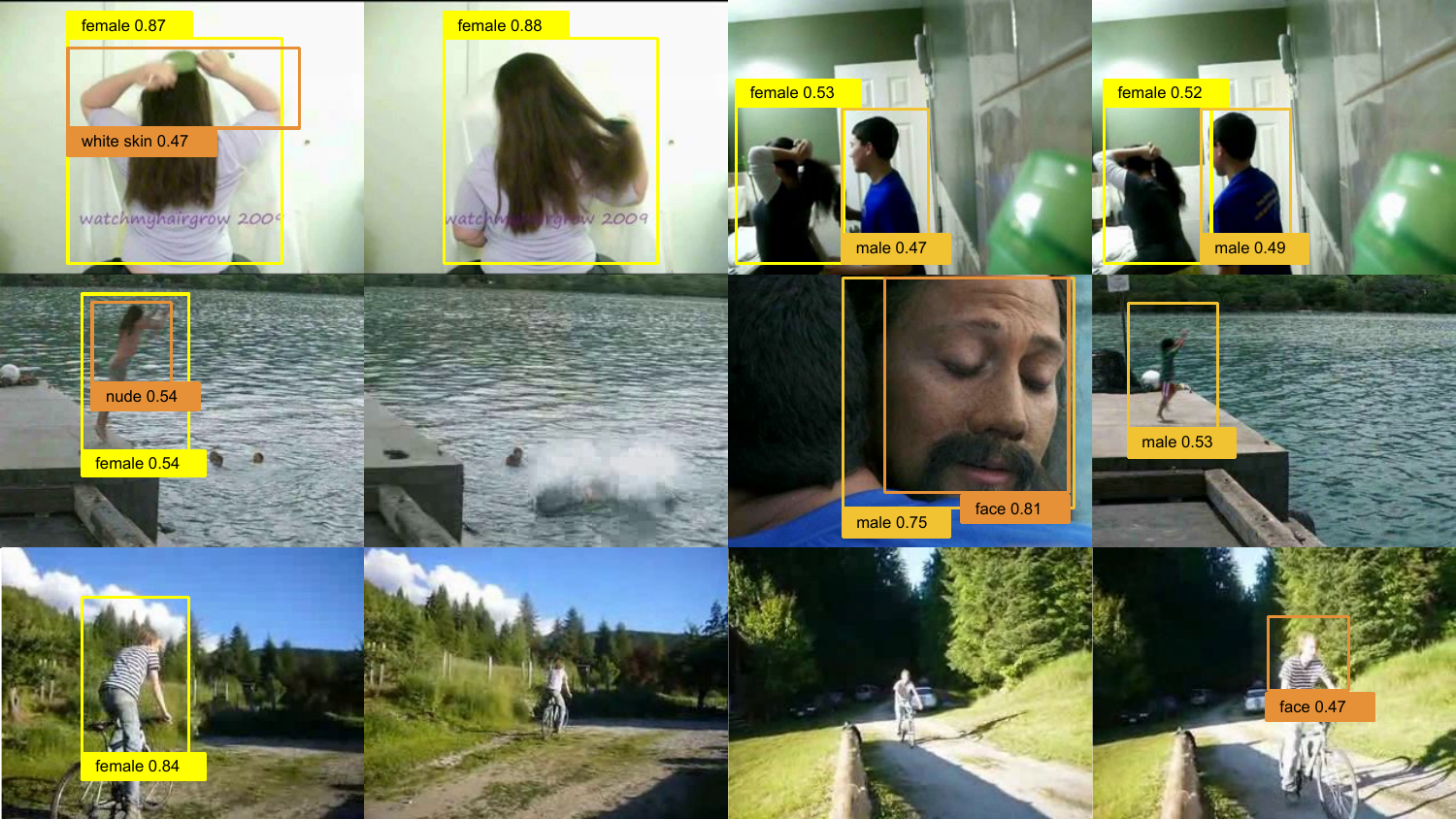}
    \caption{Object detection results on the HMDB-51 dataset.}
    \label{fig: privacy}
\end{figure}

\begin{figure}[t]
    \centering
    % \resizebox{0.7\linewidth}{0.4\linewidth}{\input{figures/plots/privacy_verif_acc}}
    \resizebox{0.7\linewidth}{0.45\linewidth}{% This file was created with tikzplotlib v0.10.1.
\begin{tikzpicture}

\definecolor{darkgray176}{RGB}{176,176,176}
\definecolor{darkorange25512714}{RGB}{255,127,14}
\definecolor{forestgreen4416044}{RGB}{44,160,44}
\definecolor{steelblue31119180}{RGB}{31,119,180}
\definecolor{lightgray204}{RGB}{204,204,204}

\begin{axis}[
legend cell align={left},
legend style={
  fill opacity=0.8,
  draw opacity=1,
  text opacity=1,
  at={(0.03,0.97)},
  anchor=north west,
  draw=lightgray204
},
tick align=outside,
tick pos=left,
% title=Probabilistic Model Checker Accuracy,
x grid style={darkgray176},
xmin=-0.025, xmax=1.025,
xtick style={color=black},
y grid style={darkgray176},
ylabel={precision},
ymin=-0.0428271619630452, ymax=1.04154333055068,
ytick style={color=black}
]

\path [draw=darkorange25512714, fill=darkorange25512714, opacity=0.2]
(axis cs:0.0,0.0260131922420807)
--(axis cs:0.0,0.00690230742577169)
--(axis cs:0.1,0.0089950735868921)
--(axis cs:0.15,0.00840845626623978)
--(axis cs:0.2,0.0158693047097012)
--(axis cs:0.25,0.0117583544109652)
--(axis cs:0.3,0.0324027896635544)
--(axis cs:0.35,0.0551597447346857)
--(axis cs:0.4,0.0889526654170129)
--(axis cs:0.45,0.15342391158137)
--(axis cs:0.5,0.324989352633913)
--(axis cs:0.55,0.581361184723236)
--(axis cs:0.6,0.779632363019299)
--(axis cs:0.65,0.906669355519591)
--(axis cs:0.7,0.953957431713891)
--(axis cs:0.75,0.943679410543571)
--(axis cs:0.8,0.967100640147221)
--(axis cs:0.85,0.965409093625469)
--(axis cs:0.9,0.956332844100327)
--(axis cs:0.95,0.972698031565311)
--(axis cs:1,0.966820496858296)
--(axis cs:1,0.989256990869366)
--(axis cs:1,0.989256990869366)
--(axis cs:0.95,0.992064811061748)
--(axis cs:0.9,0.979389504303485)
--(axis cs:0.85,0.989159097859497)
--(axis cs:0.8,0.988684344415806)
--(axis cs:0.75,0.970483066637523)
--(axis cs:0.7,0.982589518782156)
--(axis cs:0.65,0.94698560359298)
--(axis cs:0.6,0.840931565263372)
--(axis cs:0.55,0.684943165014866)
--(axis cs:0.5,0.477146924869106)
--(axis cs:0.45,0.304948219839451)
--(axis cs:0.4,0.173228966977984)
--(axis cs:0.35,0.101271823122901)
--(axis cs:0.3,0.0656981281189523)
--(axis cs:0.25,0.0382667211153158)
--(axis cs:0.2,0.0424877322844514)
--(axis cs:0.15,0.0321409079141457)
--(axis cs:0.1,0.0298468491369202)
--(axis cs:0.0,0.0260131922420807)
--cycle;

\path [draw=forestgreen4416044, fill=forestgreen4416044, opacity=0.2]
(axis cs:0.0,0.0696952313231866)
--(axis cs:0.0,0.0327943624457671)
--(axis cs:0.1,0.035224871368842)
--(axis cs:0.15,0.0587010280787703)
--(axis cs:0.2,0.0618874647569103)
--(axis cs:0.25,0.0863005157921429)
--(axis cs:0.3,0.0861201099896628)
--(axis cs:0.35,0.110303763087494)
--(axis cs:0.4,0.158101320396245)
--(axis cs:0.45,0.249475453977359)
--(axis cs:0.5,0.356815424044937)
--(axis cs:0.55,0.483905884686313)
--(axis cs:0.6,0.628882630330981)
--(axis cs:0.65,0.718333525905421)
--(axis cs:0.7,0.768947546924233)
--(axis cs:0.75,0.812059217678765)
--(axis cs:0.8,0.870783405551872)
--(axis cs:0.85,0.87787025747966)
--(axis cs:0.9,0.892830856629143)
--(axis cs:0.95,0.911940329955233)
--(axis cs:1,0.921071685612688)
--(axis cs:1,0.961616011549199)
--(axis cs:1,0.961616011549199)
--(axis cs:0.95,0.959012329448131)
--(axis cs:0.9,0.949165985343777)
--(axis cs:0.85,0.938442643787969)
--(axis cs:0.8,0.940236970142645)
--(axis cs:0.75,0.890123134045066)
--(axis cs:0.7,0.843033266313335)
--(axis cs:0.65,0.782345690561331)
--(axis cs:0.6,0.653689548668193)
--(axis cs:0.55,0.531229016134691)
--(axis cs:0.5,0.446885180655371)
--(axis cs:0.45,0.367239211112099)
--(axis cs:0.4,0.286629187887953)
--(axis cs:0.35,0.230078839343656)
--(axis cs:0.3,0.198957614038529)
--(axis cs:0.25,0.173143186514243)
--(axis cs:0.2,0.129405744405749)
--(axis cs:0.15,0.123719590175189)
--(axis cs:0.1,0.0715413858768541)
--(axis cs:0.0,0.0696952313231866)
--cycle;

\path [draw=steelblue31119180, fill=steelblue31119180, opacity=0.2]
(axis cs:0.0,0.0233022631851808)
--(axis cs:0.0,0.0064624058784876)
--(axis cs:0.1,0.0150660948282439)
--(axis cs:0.15,0.00747792813619145)
--(axis cs:0.2,0.00812410623581463)
--(axis cs:0.25,0.0170730813648771)
--(axis cs:0.3,0.0158774239986761)
--(axis cs:0.35,0.0258297581824703)
--(axis cs:0.4,0.0602547746212464)
--(axis cs:0.45,0.134564486228095)
--(axis cs:0.5,0.260256235174523)
--(axis cs:0.55,0.429000218693653)
--(axis cs:0.6,0.573574228707275)
--(axis cs:0.65,0.692103445914313)
--(axis cs:0.7,0.787981734163329)
--(axis cs:0.75,0.857848371651259)
--(axis cs:0.8,0.90882791992221)
--(axis cs:0.85,0.937630976409303)
--(axis cs:0.9,0.967515113076731)
--(axis cs:0.95,0.974337447537943)
--(axis cs:1,0.967729522908792)
--(axis cs:1,0.991268923584189)
--(axis cs:1,0.991268923584189)
--(axis cs:0.95,0.992253762709143)
--(axis cs:0.9,0.988324420188219)
--(axis cs:0.85,0.97185926448273)
--(axis cs:0.8,0.959610877342798)
--(axis cs:0.75,0.941371099938031)
--(axis cs:0.7,0.90984998986319)
--(axis cs:0.65,0.850860816134988)
--(axis cs:0.6,0.746974714240706)
--(axis cs:0.55,0.552173074716021)
--(axis cs:0.5,0.297886450538028)
--(axis cs:0.45,0.167185213576051)
--(axis cs:0.4,0.0970903069870228)
--(axis cs:0.35,0.0522817268450653)
--(axis cs:0.3,0.0389369648245967)
--(axis cs:0.25,0.039327348684409)
--(axis cs:0.2,0.0226734812676547)
--(axis cs:0.15,0.0219144502391624)
--(axis cs:0.1,0.0307549671430686)
--(axis cs:0.0,0.0233022631851808)
--cycle;

\addplot [semithick, darkorange25512714, mark=*]
table {%
0.0 0.0155286618310097
0.05 0.0155286618310097
0.1 0.0191093685470375
0.15 0.0190978891846577
0.2 0.02884062885831
0.25 0.0245557069869797
0.3 0.0489132867905463
0.35 0.0768878977699351
0.4 0.130146410610784
0.45 0.23255867251007
0.5 0.402757696398603
0.55 0.633625861341905
0.6 0.810564164035884
0.65 0.926812301648388
0.7 0.969048597256817
0.75 0.956511001010011
0.8 0.977672443734702
0.85 0.977916827312192
0.9 0.968811635554063
0.95 0.982932179718332
1 0.978583819639114
};
\addlegendentry{$\Phi_1^P$}

\addplot [semithick, forestgreen4416044, mark=*]
table {%
0.0 0.0503322605507075
0.05 0.0503322605507075
0.1 0.0527859401169337
0.15 0.0922834675204675
0.2 0.0956524445097824
0.25 0.12977081756302
0.3 0.140350874919328
0.35 0.17536404127847
0.4 0.22286391680816
0.45 0.308644872431477
0.5 0.398826600703335
0.55 0.507316086648207
0.6 0.641579275065391
0.65 0.750235097271266
0.7 0.807239272965132
0.75 0.850131965730193
0.8 0.906146764778043
0.85 0.908738560311503
0.9 0.920900780235628
0.95 0.936337522698463
1 0.941781537503254
};
\addlegendentry{$\Phi_2^P$}

\addplot [semithick, steelblue31119180, mark=*]
table {%
0 0.0144548780169523
0.05 0.0144548780169523
0.1 0.0227728229604576
0.15 0.0142584583184565
0.2 0.0151756856810615
0.25 0.0279965666613516
0.3 0.0272319409601464
0.35 0.0392574599431502
0.4 0.0788936279819318
0.45 0.150653244499782
0.5 0.279673493708256
0.55 0.489959618369621
0.6 0.661634668011808
0.65 0.766449520650113
0.7 0.850631152288143
0.75 0.897925326154069
0.8 0.934471760552729
0.85 0.954833140028798
0.9 0.978602178298043
0.95 0.983855671007872
1 0.979728604284026
};
\addlegendentry{$\Phi_3^P$}

\addplot [semithick, red, dash pattern=on 1pt off 3pt on 3pt off 3pt]
table {%
0 0
0.05 0
0.1 0
0.15 0
0.2 0
0.25 0
0.3 0
0.35 0
0.4 0
0.45 0
0.5 0
0.55 1
0.6 1
0.65 1
0.7 1
0.75 1
0.8 1
0.85 1
0.9 1
0.95 1
1 1
};
\addlegendentry{Ideal Case}
\end{axis}

\end{tikzpicture}}
    \resizebox{0.7\linewidth}{0.45\linewidth}{% This file was created with tikzplotlib v0.10.1.
\begin{tikzpicture}

\definecolor{darkgray176}{RGB}{176,176,176}
\definecolor{darkorange25512714}{RGB}{255,127,14}
\definecolor{forestgreen4416044}{RGB}{44,160,44}
\definecolor{steelblue31119180}{RGB}{31,119,180}
\definecolor{lightgray204}{RGB}{204,204,204}

\begin{axis}[
legend cell align={left},
legend style={
  fill opacity=0.8,
  draw opacity=1,
  text opacity=1,
  at={(0.03,0.39)},
  anchor=north west,
  draw=lightgray204
},
tick align=outside,
tick pos=left,
x grid style={darkgray176},
xlabel={probability returned by the model checker},
xmin=-0.025, xmax=1.025,
xtick style={color=black},
y grid style={darkgray176},
ylabel={recall},
ymin=0.0554096016006645, ymax=1.04498049516187,
ytick style={color=black}
]
\path [draw=steelblue31119180, fill=steelblue31119180, opacity=0.2]
(axis cs:0.05,1)
--(axis cs:0.05,1)
--(axis cs:0.1,0.999718981152511)
--(axis cs:0.15,0.988852475524824)
--(axis cs:0.2,0.978501362379726)
--(axis cs:0.25,0.9744467823342)
--(axis cs:0.3,0.958650647880948)
--(axis cs:0.35,0.940885143735751)
--(axis cs:0.4,0.938311020998011)
--(axis cs:0.45,0.907865039770083)
--(axis cs:0.5,0.870268369482684)
--(axis cs:0.55,0.833797699513529)
--(axis cs:0.6,0.797994906939827)
--(axis cs:0.65,0.762442559720042)
--(axis cs:0.7,0.695896698604585)
--(axis cs:0.75,0.639958946520436)
--(axis cs:0.8,0.554077064654692)
--(axis cs:0.85,0.476832272386362)
--(axis cs:0.9,0.378294440629018)
--(axis cs:0.95,0.259407701682406)
--(axis cs:1,0.134902226080617)
--(axis cs:1,0.151001177018197)
--(axis cs:1,0.151001177018197)
--(axis cs:0.95,0.289736871212014)
--(axis cs:0.9,0.417126498587077)
--(axis cs:0.85,0.525477800626737)
--(axis cs:0.8,0.605190329608429)
--(axis cs:0.75,0.679374822116829)
--(axis cs:0.7,0.736343652062695)
--(axis cs:0.65,0.788157363880694)
--(axis cs:0.6,0.833543018245555)
--(axis cs:0.55,0.874831834013068)
--(axis cs:0.5,0.902397303492661)
--(axis cs:0.45,0.934493197196383)
--(axis cs:0.4,0.966689912239267)
--(axis cs:0.35,0.990808769808835)
--(axis cs:0.3,0.998819634039841)
--(axis cs:0.25,1)
--(axis cs:0.2,1)
--(axis cs:0.15,1)
--(axis cs:0.1,1)
--(axis cs:0.05,1)
--cycle;

\path [draw=darkorange25512714, fill=darkorange25512714, opacity=0.2]
(axis cs:0.05,1)
--(axis cs:0.05,1)
--(axis cs:0.1,0.986885383817614)
--(axis cs:0.15,0.971361449258154)
--(axis cs:0.2,0.971361449258154)
--(axis cs:0.25,0.971213211656039)
--(axis cs:0.3,0.964475971799416)
--(axis cs:0.35,0.943182099050499)
--(axis cs:0.4,0.931901904174911)
--(axis cs:0.45,0.907106000032821)
--(axis cs:0.5,0.884041746485031)
--(axis cs:0.55,0.852363463556843)
--(axis cs:0.6,0.812686018387489)
--(axis cs:0.65,0.758956993734357)
--(axis cs:0.7,0.716628638410541)
--(axis cs:0.75,0.654555988235584)
--(axis cs:0.8,0.587973578034137)
--(axis cs:0.85,0.486378477759763)
--(axis cs:0.9,0.389674120377301)
--(axis cs:0.95,0.275729247839615)
--(axis cs:1,0.15476086337459)
--(axis cs:1,0.167302151188282)
--(axis cs:1,0.167302151188282)
--(axis cs:0.95,0.308686476395852)
--(axis cs:0.9,0.422475027010704)
--(axis cs:0.85,0.5366409424821)
--(axis cs:0.8,0.620906449410302)
--(axis cs:0.75,0.690182676569578)
--(axis cs:0.7,0.757104967835779)
--(axis cs:0.65,0.800763736285149)
--(axis cs:0.6,0.841193017381257)
--(axis cs:0.55,0.879378062022542)
--(axis cs:0.5,0.903349376144249)
--(axis cs:0.45,0.923151620588267)
--(axis cs:0.4,0.941179878811092)
--(axis cs:0.35,0.957079872949364)
--(axis cs:0.3,0.968922793651887)
--(axis cs:0.25,0.98743363043881)
--(axis cs:0.2,0.98743363043881)
--(axis cs:0.15,0.99021952368575)
--(axis cs:0.1,0.994672669363792)
--(axis cs:0.05,1)
--cycle;

\path [draw=forestgreen4416044, fill=forestgreen4416044, opacity=0.2]
(axis cs:0.05,1)
--(axis cs:0.05,1)
--(axis cs:0.1,0.998019433514293)
--(axis cs:0.15,0.998019433514293)
--(axis cs:0.2,0.980645916985713)
--(axis cs:0.25,0.968405779015307)
--(axis cs:0.3,0.96082044060075)
--(axis cs:0.35,0.938614240956674)
--(axis cs:0.4,0.91610737671658)
--(axis cs:0.45,0.878542771391973)
--(axis cs:0.5,0.848249964552917)
--(axis cs:0.55,0.79653220642805)
--(axis cs:0.6,0.741516552163964)
--(axis cs:0.65,0.6811074794401)
--(axis cs:0.7,0.615932470282028)
--(axis cs:0.75,0.545753866598249)
--(axis cs:0.8,0.465942002297824)
--(axis cs:0.85,0.382452144318257)
--(axis cs:0.9,0.301690544522067)
--(axis cs:0.95,0.20920123581874)
--(axis cs:1,0.100390096762538)
--(axis cs:1,0.117989415584609)
--(axis cs:1,0.117989415584609)
--(axis cs:0.95,0.222558995362946)
--(axis cs:0.9,0.325244896758354)
--(axis cs:0.85,0.413273291079261)
--(axis cs:0.8,0.50208544264476)
--(axis cs:0.75,0.576172659230347)
--(axis cs:0.7,0.652566683477286)
--(axis cs:0.65,0.709753874734217)
--(axis cs:0.6,0.771822576732982)
--(axis cs:0.55,0.816848451408925)
--(axis cs:0.5,0.864547890340991)
--(axis cs:0.45,0.904040281050814)
--(axis cs:0.4,0.926633139369971)
--(axis cs:0.35,0.946910113731304)
--(axis cs:0.3,0.968241238468976)
--(axis cs:0.25,0.984023115278707)
--(axis cs:0.2,0.998589943188836)
--(axis cs:0.15,1)
--(axis cs:0.1,0.99949486042539)
--(axis cs:0.05,1)
--cycle;

\addplot [semithick, darkorange25512714, mark=*]
table {%
0.05 1
0.1 0.989775674608715
0.15 0.983004867794238
0.2 0.980224923300644
0.25 0.978504509558323
0.3 0.96698265137985
0.35 0.951934721881483
0.4 0.936580607559105
0.45 0.913066694802792
0.5 0.892858614108847
0.55 0.86663391318374
0.6 0.831147619006838
0.65 0.786671300951885
0.7 0.742971157572613
0.75 0.677165944004006
0.8 0.609764003467827
0.85 0.512592342608368
0.9 0.402721660812571
0.95 0.289803936219155
1 0.159864531212353
};
\addlegendentry{$\Phi_1^P$}

\addplot [semithick, forestgreen4416044, mark=*]
table {%
0.05 1
0.1 0.998834671596821
0.15 0.998834671596821
0.2 0.991333410487319
0.25 0.976922554446781
0.3 0.964831654433394
0.35 0.941825757924165
0.4 0.919987542024427
0.45 0.887466331429979
0.5 0.855505775200185
0.55 0.807670918431496
0.6 0.75580957844921
0.65 0.694724727389608
0.7 0.633050936471532
0.75 0.564466952463761
0.8 0.486302813310428
0.85 0.399018133121221
0.9 0.31339246348747
0.95 0.217249045423869
1 0.111701018521459
};
\addlegendentry{$\Phi_2^P$}

\addplot [semithick, steelblue31119180, mark=*]
table {%
0.05 1
0.1 0.999906327050837
0.15 0.996284158508275
0.2 0.992833787459909
0.25 0.989327549133181
0.3 0.983668715662044
0.35 0.970581318692347
0.4 0.955737311402506
0.45 0.924911144165581
0.5 0.888917890700446
0.55 0.856430178432109
0.6 0.820371653325498
0.65 0.776256764915082
0.7 0.711649125063264
0.75 0.65477494648894
0.8 0.578442136231656
0.85 0.497512426003029
0.9 0.393967248939646
0.95 0.269587310542402
1 0.14294571094026
};
\addlegendentry{$\Phi_3^P$}

\addplot [semithick, red, dash pattern=on 1pt off 3pt on 3pt off 3pt]
table {%
0 1
0.05 1
0.1 1
0.15 1
0.2 1
0.25 1
0.3 1
0.35 1
0.4 1
0.45 1
0.5 1
0.55 1
0.6 1
0.65 1
0.7 1
0.75 1
0.8 1
0.85 1
0.9 1
0.95 1
1 1
};
\addlegendentry{Ideal Case}
\end{axis}

\end{tikzpicture}}
    \caption{The top and bottom figures show the verification precision and recall of verification results whose probabilities fall into each particular range.}
    \label{fig: privacy-verif}
\end{figure}
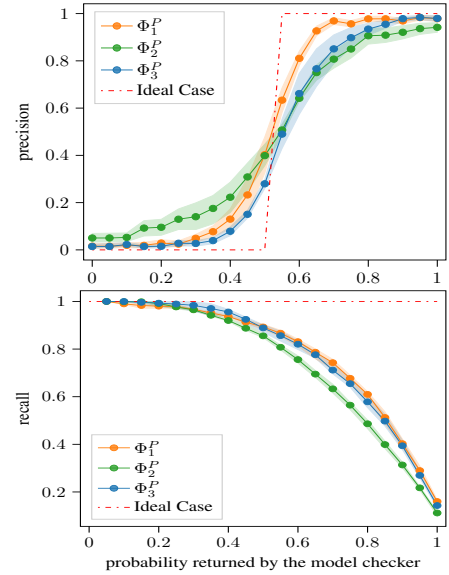

We apply the proposed method on a realistic video dataset HMDB-51 \cite{HMDB} to verify the videos against several privacy rules and find all the videos that violate any of the privacy rules.
We evaluate the proposed method in the metrics defined in Definition \ref{def: metrics}.
\begin{definition}
    \label{def: metrics}
    Let $\Phi$ be an \gls{tl} specification, $N$ be the number of videos whose verification probabilities returned by the model checker are between an interval $[c_1, c_2)$, $T_P$ be the number of videos whose verification probabilities are between $[c_1, c_2)$ that actually satisfy $\Phi$, $T_P*$ be the number of videos whose verification probabilities are between $[c_1, 1]$ that satisfy $\Phi$, $A_P$ be the total number of videos in the dataset that satisfy $\Phi$,
    we define \textsc{precision} $P_c$ and \textsc{recall} $R_c$ as
    \begin{center}
        $P_c = \frac{T_P}{N}$ and $R_c = \frac{T_P*}{A_P}$.
    \end{center}
    In the experiments, we equally divide the probabilities into 20 intervals: $[0, 0.05), [0.05, 0.1),..., [0.9, 0.95), [0.95,1]$.
\end{definition}

In this setting, we obtain privacy annotations of the HMDB-51 dataset from PA-HMDB51 \cite{wang2019privacy}, which include characters' genders, races, etc., from the videos. We use these annotations as the ground truth to compute the accuracies.

As for the first step, we take some of the annotations from PA-HMDB51 as the propositions and build a set of privacy rules by querying GPT-4:
\vspace{0.2cm}
\begin{lstlisting}[language=completion]
    Given the propositions: female, male, face, nude, black skin, white skin.
    Transform the following rules into temporal logic formulas:
    1. Never reveal gender.
    2. Don't show the faces of nude people.
    3. Never reveal races.
\end{lstlisting}
\vspace{0.2cm}
We then transform these rules above into \gls{tl} specifications:

$\Phi_1^P = \lalways ( \neg \text{male } \land \neg$ female),
    
$\Phi_2^P = \lalways$ (nude $\rightarrow \neg$ face),
    
$\Phi_3^P = \lalways ( \neg \text{black skin } \land \neg$ white skin).

We extract all the frames from videos and apply the Grounded-SAM to detect whether the objects described in the propositions appear in each frame.
We obtain the confidence scores from the Grounded-SAM and follow Algorithm \ref{alg: frames2automata} to construct a probabilistic automaton for each video.
The privacy annotations from the dataset are static scenes. Hence, we only consider the detection results from the \gls{vlm} when constructing the automaton.
Next, we use a probabilistic model checker to compute the probability that each video satisfies the \gls{tl} specifications. By doing so, we can efficiently find all videos that violate any of the specifications, i.e., verification probability below 50\%.

We evaluate our verification results (probabilities) using the ground truth privacy annotations from the dataset.
We only select the following annotations to avoid ambiguity: race-black, race-white, gender-female, gender-male, gender-coexist, and nudity-semi-nudity.
We evaluate the verification results over the metrics in Definition \ref{def: metrics}. For each interval, we compute the precision and recall, as presented in Figure \ref{fig: privacy-verif}.
If we use 0.5 as the threshold to determine whether we should add the video to the search result, the precision is approximately 90\%, and the recall is over 80\%, indicating the reliability of our verification results.

\paragraph{Search over Autonomous Driving Videos.}
\begin{figure}[t]
    \centering
    \includegraphics[width=\linewidth]{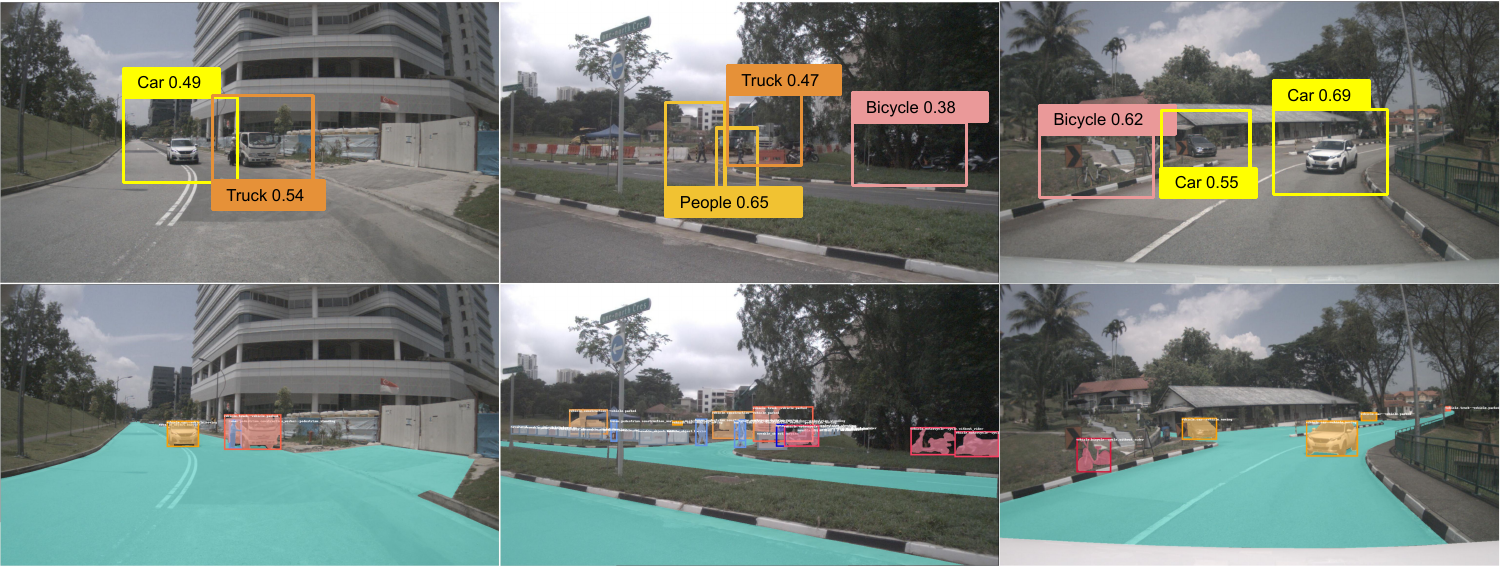}
    \caption{Examples of the object detection results on images from the NuScenes Dataset. The top row shows the object detection results by the Grounded-SAM, and the bottom row shows the annotated segments provided by the dataset.}
    \label{fig: nuimages}
\end{figure}

\begin{figure}[t]
    \centering
    % \resizebox{0.7\linewidth}{0.4\linewidth}{\input{figures/plots/traffic_acc}}
    \resizebox{0.7\linewidth}{0.45\linewidth}{% This file was created with tikzplotlib v0.10.1.
\begin{tikzpicture}

\definecolor{darkgray176}{RGB}{176,176,176}
\definecolor{darkorange25512714}{RGB}{255,127,14}
\definecolor{forestgreen4416044}{RGB}{44,160,44}
\definecolor{steelblue31119180}{RGB}{31,119,180}

\definecolor{lightgray204}{RGB}{204,204,204}

\begin{axis}[
legend cell align={left},
legend style={
  fill opacity=0.8,
  draw opacity=1,
  text opacity=1,
  at={(0.03,0.97)},
  anchor=north west,
  draw=lightgray204
},
tick align=outside,
tick pos=left,
x grid style={darkgray176},
xmin=-0.025, xmax=1.0475,
xtick style={color=black},
y grid style={darkgray176},
ylabel={precision},
ymin=-0.05, ymax=1.05,
ytick style={color=black}
]
\path [draw=steelblue31119180, fill=steelblue31119180, opacity=0.2]
(axis cs:0.05,0.0253636453868406)
--(axis cs:0.05,0.00305413917471126)
--(axis cs:0.1,0)
--(axis cs:0.15,0)
--(axis cs:0.2,0)
--(axis cs:0.25,0.0035399045418395)
--(axis cs:0.3,0.024327700892027)
--(axis cs:0.35,0.0290286047412729)
--(axis cs:0.4,0.121730603554812)
--(axis cs:0.45,0.283320594095879)
--(axis cs:0.5,0.491030442700137)
--(axis cs:0.55,0.73403309172525)
--(axis cs:0.6,0.920385016065273)
--(axis cs:0.65,0.945635626802159)
--(axis cs:0.7,0.968854475794632)
--(axis cs:0.75,0.972051127113263)
--(axis cs:0.8,0.957572735084183)
--(axis cs:0.85,0.987030590862908)
--(axis cs:0.9,0.945715480599243)
--(axis cs:0.95,0.95040133622876)
--(axis cs:1,0.97525620521459)
--(axis cs:1,0.998246925012311)
--(axis cs:1,0.998246925012311)
--(axis cs:0.95,0.998260697511306)
--(axis cs:0.9,0.981162845383782)
--(axis cs:0.85,1)
--(axis cs:0.8,0.990744705663867)
--(axis cs:0.75,1)
--(axis cs:0.7,0.984142645862291)
--(axis cs:0.65,1)
--(axis cs:0.6,0.983878724047964)
--(axis cs:0.55,0.909277070028577)
--(axis cs:0.5,0.780985971067399)
--(axis cs:0.45,0.619066517373452)
--(axis cs:0.4,0.389207077558803)
--(axis cs:0.35,0.189980985522293)
--(axis cs:0.3,0.0754635937868059)
--(axis cs:0.25,0.0156859957465823)
--(axis cs:0.2,0.0255074102646862)
--(axis cs:0.15,0.0204572954419358)
--(axis cs:0.1,0.0248800336675582)
--(axis cs:0.05,0.0253636453868406)
--cycle;

\path [draw=darkorange25512714, fill=darkorange25512714, opacity=0.2]
(axis cs:0.05,0.0225223425288228)
--(axis cs:0.05,0)
--(axis cs:0.1,0)
--(axis cs:0.15,0)
--(axis cs:0.2,0.0153734911336542)
--(axis cs:0.25,0)
--(axis cs:0.3,0.00684511820786916)
--(axis cs:0.35,0.0402825259490455)
--(axis cs:0.4,0.0978347645189898)
--(axis cs:0.45,0.221882077608977)
--(axis cs:0.5,0.570015991536628)
--(axis cs:0.55,0.814602188033576)
--(axis cs:0.6,0.929222449842715)
--(axis cs:0.65,0.966494781373303)
--(axis cs:0.7,0.977470317144192)
--(axis cs:0.75,0.966675818567341)
--(axis cs:0.8,0.993100155075998)
--(axis cs:0.85,0.978751465940861)
--(axis cs:0.9,0.977462402434298)
--(axis cs:0.95,0.989373058641834)
--(axis cs:1,1)
--(axis cs:1,1)
--(axis cs:1,1)
--(axis cs:0.95,0.998474261202331)
--(axis cs:0.9,0.997236803483988)
--(axis cs:0.85,1)
--(axis cs:0.8,1)
--(axis cs:0.75,0.993422555320223)
--(axis cs:0.7,0.994925762660479)
--(axis cs:0.65,0.995922351414908)
--(axis cs:0.6,0.996058084609907)
--(axis cs:0.55,0.974534216334374)
--(axis cs:0.5,0.803150786241199)
--(axis cs:0.45,0.495729617578656)
--(axis cs:0.4,0.212528674256623)
--(axis cs:0.35,0.083342825874132)
--(axis cs:0.3,0.0406931445991213)
--(axis cs:0.25,0.00850213471131216)
--(axis cs:0.2,0.0328865033694277)
--(axis cs:0.15,0.0200556086101762)
--(axis cs:0.1,0.022533278532657)
--(axis cs:0.05,0.0225223425288228)
--cycle;

\path [draw=forestgreen4416044, fill=forestgreen4416044, opacity=0.2]
(axis cs:0.05,0.0421494967755373)
--(axis cs:0.05,0.0016442143889907)
--(axis cs:0.1,0)
--(axis cs:0.15,0.0319592919075119)
--(axis cs:0.2,0)
--(axis cs:0.25,0.0544060682545722)
--(axis cs:0.3,0.0326409328601846)
--(axis cs:0.35,0.0501185811418294)
--(axis cs:0.4,0.095621148003067)
--(axis cs:0.45,0.211648578307941)
--(axis cs:0.5,0.375620068734824)
--(axis cs:0.55,0.553342838617992)
--(axis cs:0.6,0.679725246428011)
--(axis cs:0.65,0.80552892453881)
--(axis cs:0.7,0.874082352935172)
--(axis cs:0.75,0.908450971249317)
--(axis cs:0.8,0.970513622355816)
--(axis cs:0.85,0.988852855378028)
--(axis cs:0.9,0.970327343305148)
--(axis cs:0.95,0.928777899807093)
--(axis cs:1,0.934726077241744)
--(axis cs:1,0.997324767083776)
--(axis cs:1,0.997324767083776)
--(axis cs:0.95,1)
--(axis cs:0.9,1)
--(axis cs:0.85,1)
--(axis cs:0.8,0.993780364515538)
--(axis cs:0.75,0.974278105164873)
--(axis cs:0.7,0.972298613028078)
--(axis cs:0.65,0.91613205243375)
--(axis cs:0.6,0.814883442129719)
--(axis cs:0.55,0.704775450798365)
--(axis cs:0.5,0.570795181392099)
--(axis cs:0.45,0.429428012757189)
--(axis cs:0.4,0.329785899891441)
--(axis cs:0.35,0.184431191186007)
--(axis cs:0.3,0.107248499144856)
--(axis cs:0.25,0.108509192322081)
--(axis cs:0.2,0.0293330213007519)
--(axis cs:0.15,0.0493716944780972)
--(axis cs:0.1,0.0177546892587035)
--(axis cs:0.05,0.0421494967755373)
--cycle;

\addplot [semithick, steelblue31119180, mark=*]
table {%
0.05 0.0143281751438148
0.1 0.0111572583666989
0.15 0.00681909848064526
0.2 0.0100669976680643
0.25 0.00958689887243588
0.3 0.0483338463747917
0.35 0.108867140048254
0.4 0.250266602729035
0.45 0.446763083503244
0.5 0.63738074331841
0.55 0.82089516573679
0.6 0.954577442308443
0.65 0.979977304133951
0.7 0.976539146156413
0.75 0.989648658656735
0.8 0.973889542940845
0.85 0.993577640391707
0.9 0.963962629610269
0.95 0.975392633711913
1 0.98675156511345
};
\addlegendentry{$\Phi_1^T$}
\addplot [semithick, darkorange25512714, mark=*]
table {%
0.05 0.0101053337864178
0.1 0.007511092844219
0.15 0.00668520287005872
0.2 0.0237701958486283
0.25 0.00283404490377072
0.3 0.0250941906768199
0.35 0.0573968105512212
0.4 0.153856089408976
0.45 0.364000232047867
0.5 0.691412882248203
0.55 0.894568202183975
0.6 0.963825872293834
0.65 0.980656462483089
0.7 0.985698968348316
0.75 0.979897470058805
0.8 0.997066082232178
0.85 0.990752631023881
0.9 0.987570091478717
0.95 0.994185063837689
1 1
};
\addlegendentry{$\Phi_2^T$}
\addplot [semithick, forestgreen4416044, mark=*]
table {%
0.05 0.0194457436300374
0.1 0.00622983725839842
0.15 0.0418134145870199
0.2 0.0097776737669173
0.25 0.0752515790411059
0.3 0.0595491096373276
0.35 0.114149218765852
0.4 0.199118198417281
0.45 0.312802239354157
0.5 0.466745004338895
0.55 0.651235925814316
0.6 0.748731478080881
0.65 0.863861158873341
0.7 0.922810075481022
0.75 0.94176726572694
0.8 0.982590345943559
0.85 0.996284285126009
0.9 0.990109114435049
0.95 0.971487644388498
1 0.974843512548842
};
\addlegendentry{$\Phi_3^T$}

\addplot [semithick, red, dash pattern=on 1pt off 3pt on 3pt off 3pt]
table {%
0 0
0.05 0
0.1 0
0.15 0
0.2 0
0.25 0
0.3 0
0.35 0
0.4 0
0.45 0
0.5 0
0.55 1
0.6 1
0.65 1
0.7 1
0.75 1
0.8 1
0.85 1
0.9 1
0.95 1
1 1
};
\addlegendentry{Ideal Case}
\end{axis}

\end{tikzpicture}}
    \resizebox{0.7\linewidth}{0.45\linewidth}{% This file was created with tikzplotlib v0.10.1.
\begin{tikzpicture}

\definecolor{darkgray176}{RGB}{176,176,176}
\definecolor{darkorange25512714}{RGB}{255,127,14}
\definecolor{forestgreen4416044}{RGB}{44,160,44}
\definecolor{steelblue31119180}{RGB}{31,119,180}

\definecolor{lightgray204}{RGB}{204,204,204}

\begin{axis}[
legend cell align={left},
legend style={
  fill opacity=0.8,
  draw opacity=1,
  text opacity=1,
  at={(0.03,0.39)},
  anchor=north west,
  draw=lightgray204
},
tick align=outside,
tick pos=left,
x grid style={darkgray176},
xlabel={probability returned by the model checker},
xmin=-0.025, xmax=1.0475,
xtick style={color=black},
y grid style={darkgray176},
ylabel={recall},
ymin=0.0467361703067443, ymax=1.04539351569968,
ytick style={color=black}
]
\path [draw=steelblue31119180, fill=steelblue31119180, opacity=0.2]
(axis cs:0.05,1)
--(axis cs:0.05,1)
--(axis cs:0.1,0.993764573604331)
--(axis cs:0.15,0.988540162895121)
--(axis cs:0.2,0.977314289639379)
--(axis cs:0.25,0.962651671769852)
--(axis cs:0.3,0.951219104535251)
--(axis cs:0.35,0.931822976622635)
--(axis cs:0.4,0.91748178570699)
--(axis cs:0.45,0.896347091280172)
--(axis cs:0.5,0.868282862867662)
--(axis cs:0.55,0.847136691717279)
--(axis cs:0.6,0.795025949339769)
--(axis cs:0.65,0.742804119462223)
--(axis cs:0.7,0.68312879950591)
--(axis cs:0.75,0.617093258673893)
--(axis cs:0.8,0.54703411983055)
--(axis cs:0.85,0.448207285266098)
--(axis cs:0.9,0.350537510061891)
--(axis cs:0.95,0.242445909044007)
--(axis cs:1,0.126218642572315)
--(axis cs:1,0.13493082306486)
--(axis cs:1,0.13493082306486)
--(axis cs:0.95,0.258836087599586)
--(axis cs:0.9,0.377322619811952)
--(axis cs:0.85,0.476370559676327)
--(axis cs:0.8,0.565149079945154)
--(axis cs:0.75,0.649866997880398)
--(axis cs:0.7,0.711358176517771)
--(axis cs:0.65,0.764371687417778)
--(axis cs:0.6,0.818642353757695)
--(axis cs:0.55,0.857505873674379)
--(axis cs:0.5,0.89294497044052)
--(axis cs:0.45,0.920923550511601)
--(axis cs:0.4,0.954331435950199)
--(axis cs:0.35,0.962658226632852)
--(axis cs:0.3,0.979212281738612)
--(axis cs:0.25,0.992356576153454)
--(axis cs:0.2,0.993946451044242)
--(axis cs:0.15,0.997076974451468)
--(axis cs:0.1,1)
--(axis cs:0.05,1)
--cycle;

\path [draw=darkorange25512714, fill=darkorange25512714, opacity=0.2]
(axis cs:0.05,1)
--(axis cs:0.05,1)
--(axis cs:0.1,0.990943470927163)
--(axis cs:0.15,0.982798094885411)
--(axis cs:0.2,0.977015657698914)
--(axis cs:0.25,0.974713274529544)
--(axis cs:0.3,0.967020252067543)
--(axis cs:0.35,0.949601948135603)
--(axis cs:0.4,0.942466138544476)
--(axis cs:0.45,0.929312900534251)
--(axis cs:0.5,0.912629004778893)
--(axis cs:0.55,0.877908416439097)
--(axis cs:0.6,0.843258052557992)
--(axis cs:0.65,0.810806551579494)
--(axis cs:0.7,0.751648641024557)
--(axis cs:0.75,0.697314081754056)
--(axis cs:0.8,0.618143752827407)
--(axis cs:0.85,0.520835113157711)
--(axis cs:0.9,0.422894524925313)
--(axis cs:0.95,0.304864118701696)
--(axis cs:1,0.160142543977571)
--(axis cs:1,0.1707018768563)
--(axis cs:1,0.1707018768563)
--(axis cs:0.95,0.317859489116645)
--(axis cs:0.9,0.441568983333159)
--(axis cs:0.85,0.549218352457187)
--(axis cs:0.8,0.653288691626303)
--(axis cs:0.75,0.722671670145478)
--(axis cs:0.7,0.789661819985433)
--(axis cs:0.65,0.844472591081691)
--(axis cs:0.6,0.878496775882344)
--(axis cs:0.55,0.910826163368695)
--(axis cs:0.5,0.930116593476596)
--(axis cs:0.45,0.946853393825295)
--(axis cs:0.4,0.96715466329955)
--(axis cs:0.35,0.974595316432881)
--(axis cs:0.3,0.986889426663325)
--(axis cs:0.25,0.994205973482603)
--(axis cs:0.2,0.996255828977761)
--(axis cs:0.15,0.999332443171132)
--(axis cs:0.1,0.999332443171132)
--(axis cs:0.05,1)
--cycle;

\path [draw=forestgreen4416044, fill=forestgreen4416044, opacity=0.2]
(axis cs:0.05,1)
--(axis cs:0.05,1)
--(axis cs:0.1,0.987022144620748)
--(axis cs:0.15,0.977547111985152)
--(axis cs:0.2,0.965170850230299)
--(axis cs:0.25,0.941823169757731)
--(axis cs:0.3,0.914754789302595)
--(axis cs:0.35,0.895108109576169)
--(axis cs:0.4,0.863516484871588)
--(axis cs:0.45,0.826971326297811)
--(axis cs:0.5,0.779611339243891)
--(axis cs:0.55,0.733111414836004)
--(axis cs:0.6,0.680522485996098)
--(axis cs:0.65,0.633083268218493)
--(axis cs:0.7,0.568190284628592)
--(axis cs:0.75,0.497102762820809)
--(axis cs:0.8,0.433873859930154)
--(axis cs:0.85,0.354012963663661)
--(axis cs:0.9,0.271065822850027)
--(axis cs:0.95,0.179424521185104)
--(axis cs:1,0.0921296860064232)
--(axis cs:1,0.10759411377203)
--(axis cs:1,0.10759411377203)
--(axis cs:0.95,0.213125457470732)
--(axis cs:0.9,0.307148376970599)
--(axis cs:0.85,0.401291913121103)
--(axis cs:0.8,0.483740161856525)
--(axis cs:0.75,0.547537316973402)
--(axis cs:0.7,0.620708885933941)
--(axis cs:0.65,0.678358626715675)
--(axis cs:0.6,0.736778544485701)
--(axis cs:0.55,0.789989181756581)
--(axis cs:0.5,0.836050317826918)
--(axis cs:0.45,0.87487453286751)
--(axis cs:0.4,0.907801307817508)
--(axis cs:0.35,0.931998377195175)
--(axis cs:0.3,0.949592279883137)
--(axis cs:0.25,0.960478174682455)
--(axis cs:0.2,0.979227857838248)
--(axis cs:0.15,0.993942838258156)
--(axis cs:0.1,0.997626660838039)
--(axis cs:0.05,1)
--cycle;

\addplot [semithick, steelblue31119180, mark=*]
table {%
0.05 1
0.1 0.997352226244212
0.15 0.993450890144516
0.2 0.985663973573058
0.25 0.979732719172058
0.3 0.965635401033509
0.35 0.949072250244845
0.4 0.935393897438784
0.45 0.906460505107357
0.5 0.87970335569574
0.55 0.851622052240397
0.6 0.807324982503262
0.65 0.75358790344
0.7 0.696218085485926
0.75 0.630688075689498
0.8 0.556508984192744
0.85 0.463778236539858
0.9 0.363717143107977
0.95 0.250601741744862
1 0.130568732929548
};
\addlegendentry{$\Phi_1^T$}
\addplot [semithick, darkorange25512714, mark=*]
table {%
0.05 1
0.1 0.995137957049148
0.15 0.992511905386555
0.2 0.98762915967695
0.25 0.985326776507579
0.3 0.978233047635141
0.35 0.961864385059187
0.4 0.954657867689037
0.45 0.937826148469172
0.5 0.922459027526193
0.55 0.894367289903896
0.6 0.861945645019624
0.65 0.826431282857774
0.7 0.770655230504995
0.75 0.709371301765978
0.8 0.636502823499829
0.85 0.536477179596415
0.9 0.432231754129236
0.95 0.311361803909171
1 0.165802386454667
};
\addlegendentry{$\Phi_2^T$}
\addplot [semithick, forestgreen4416044, mark=*]
table {%
0.05 1
0.1 0.99224936561682
0.15 0.985764731551617
0.2 0.972176130147276
0.25 0.95195111339974
0.3 0.932295964209767
0.35 0.913553243385672
0.4 0.887254594942144
0.45 0.852974517372877
0.5 0.810935233637448
0.55 0.763396404858173
0.6 0.709316382155125
0.65 0.656497153317061
0.7 0.596615869445898
0.75 0.523869852459421
0.8 0.45738227959575
0.85 0.376221423846002
0.9 0.288341950908986
0.95 0.196298015812242
1 0.099807187856689
};
\addlegendentry{$\Phi_3^T$}

\addplot [semithick, red, dash pattern=on 1pt off 3pt on 3pt off 3pt]
table {%
0 1
0.05 1
0.1 1
0.15 1
0.2 1
0.25 1
0.3 1
0.35 1
0.4 1
0.45 1
0.5 1
0.55 1
0.6 1
0.65 1
0.7 1
0.75 1
0.8 1
0.85 1
0.9 1
0.95 1
1 1
};
\addlegendentry{Ideal Case}
\end{axis}

\end{tikzpicture}}
    \caption{The top and bottom figures show the verification precision and recall of verification results whose probabilities fall into each range.}
    \label{fig: traffic-verif}
\end{figure}
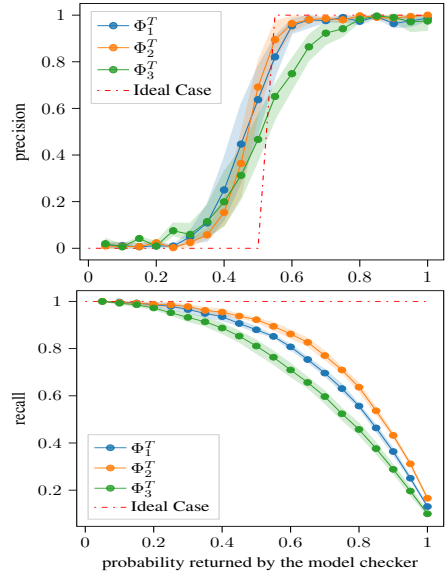

We use the proposed method to search videos that satisfy the provided specifications within the NuScenes Dataset \cite{nuimages}. The NuScenes Dataset consists of 93,000 images with annotations of objects. We group 10 images and consider them as frames for one video clip. Then, we want to search for the video clips that satisfy our manually generated specifications:

$\Phi_1^T = \lalways$ bicycle $\rightarrow$ human,

$\Phi_2^T = \lalways$ truck $\rightarrow (\leventually$ car),

$\Phi_3^T = \leventually$ car $\land (\lnext \neg$ human ).

From the specifications, we extract a set of propositions $\{$ bicycle, human, truck, car$\}$. We apply the Grounded-SAM to detect objects described by these propositions and follow Algorithm \ref{alg: frames2automata} to construct a probabilistic automaton for each 10-frame video clip. Then, we use the probabilistic model checker to compute the probabilities of each video clip satisfying each specification. If a video satisfies all three specifications with a probability above 50\%, we extract it as one of the search results.

Since the dataset contains annotations of objects described by the propositions, we take these annotations to evaluate our verification results.
We present both the object detection results and the annotations provided by the dataset in Figure \ref{fig: nuimages}.
We evaluate the probabilistic verification results over the metrics in Definition \ref{def: metrics}. 
We indicate the reliability of the method in Figure \ref{fig: traffic-verif}. The precision of verification results whose probabilities are greater than 0.5 is around 95\%, and the recall is above 80\%.
Although there exist small gaps between the empirical results and the ideal case, these gaps can be filled by the future development of \glspl{vlm}.

\section{Conclusions}

We design a method for efficient video search, which consists of an algorithm that maps text-based event descriptions into LTL$_f$ formulas and an algorithm for constructing a probabilistic automaton encoding the video information. We calibrate the confidence and accuracy of the vision and language foundation model and use the calibrated confidence as probabilities in the probabilistic automaton. Hence, we can apply formal verification to compute the probability that the automaton of each video satisfies the LTL$_f$ formulas. By doing so, we can efficiently find all the videos that satisfy the event description with formal guarantees, i.e., probabilities above a certain threshold. Due to the limitation of the selected foundation model, our method only works for static event descriptions. As a future direction, we can incorporate action prediction models for searching dynamic events.

% \newpage
% \bibliographystyle{aaai24}
\bibliography{references}

\newpage
\appendix
\setcounter{secnumdepth}{2}
\section{Additional Preliminaries}
\subsection{Probabilistic Verification}
Probabilistic verification is a procedure of verifying a finite state machine (e.g., probabilistic automaton) against some provided specifications and returning a score indicating the probability of the specifications being satisfied.

In practice, each trajectory from the probabilistic automaton is associated with a probability and may or may not satisfy the specification. The model checker sums up the probabilities of all the trajectories that satisfy the specification and obtains a probability that the automaton satisfies the specification.

We use the College Introductory Videos as an example: According to the specification $\Phi$, we expect the label of every state in the trajectory to be ``faces = False." 
Hence, there is only one trajectory in each automaton that will satisfy $\Phi$, whose states only consist of the label ``faces = False." 

\begin{figure}[t]
    \centering
    \includegraphics[width=\linewidth]{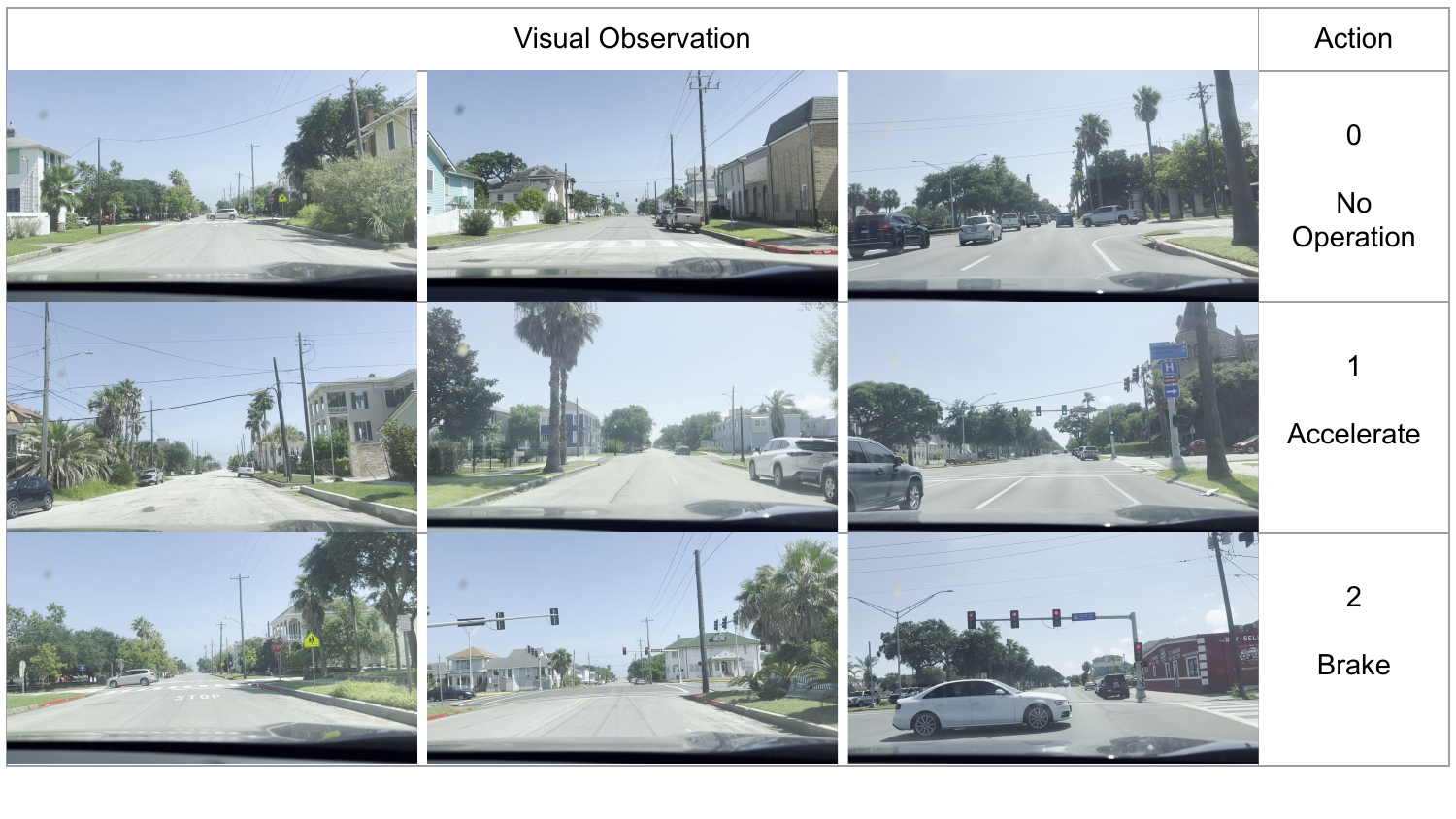}
    \caption{Sample video frames and their labels from the Driving Control Dataset.}
    \label{fig: dataset}
\end{figure}

\begin{figure}
    \centering
    \resizebox{0.8\linewidth}{0.6\linewidth}{% This file was created with tikzplotlib v0.10.1.
\begin{tikzpicture}

\definecolor{darkgray176}{RGB}{176,176,176}
\definecolor{darkorange25512714}{RGB}{255,127,14}
\definecolor{forestgreen4416044}{RGB}{44,160,44}
\definecolor{steelblue31119180}{RGB}{31,119,180}
\definecolor{lightgray204}{RGB}{204,204,204}

\begin{axis}[
legend cell align={left},
legend style={
  fill opacity=0.8,
  draw opacity=1,
  text opacity=1,
  at={(0.85,0.3)},
  anchor=north west,
  draw=lightgray204
},
% title=Probabilistic Verification on Each Chunk of Frames,
tick align=outside,
tick pos=left,
x grid style={darkgray176},
xlabel={chunk number},
xmin=0.55, xmax=10.45,
xtick style={color=black},
y grid style={darkgray176},
ylabel={probability of satisfying $\Phi$},
ymin=0.307, ymax=1.033,
ytick style={color=black}
]

\addplot [draw=white, fill=steelblue31119180, mark=*, only marks]
table{%
x  y
1 0.42
2 0.78
3 0.51
4 1
5 1
6 1
7 1
8 1
9 0.67
10 0.81
};

\addplot [draw=white, fill=darkorange25512714, mark=*, only marks]
table{%
x  y
1 1
2 1
3 1
4 1
5 1
6 1
7 1
8 0.34
9 1
10 1
};

\addplot [draw=white, fill=forestgreen4416044, mark=*, only marks]
table{%
x  y
1 1
2 1
3 1
4 1
5 1
6 1
7 1
8 1
9 1
10 1
};
\addplot [semithick, steelblue31119180, dash pattern=on 1pt off 3pt on 3pt off 3pt]
table {%
1 0.42
2 0.78
3 0.51
4 1
5 1
6 1
7 1
8 1
9 0.67
10 0.81
};
\addlegendentry{$\Phi_2$}
\addplot [semithick, darkorange25512714]
table {%
1 1
2 1
3 1
4 1
5 1
6 1
7 1
8 0.34
9 1
10 1
};
\addlegendentry{$\Phi_3$}
\addplot [semithick, forestgreen4416044, dotted]
table {%
1 1
2 1
3 1
4 1
5 1
6 1
7 1
8 1
9 1
10 1
};
\addlegendentry{$\Phi_1$}
\end{axis}

\end{tikzpicture}}
    \caption{Probability of each chunk of frames (30 frames) from the Driving Control Dataset satisfying each of the provided specifications (traffic rules). We record the probabilities of 10 chunks.}
    \label{fig: drive-spec}
\end{figure}
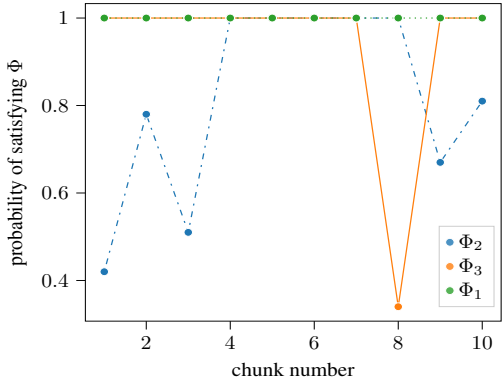

\section{Driving Control Dataset}
\label{sec: dataset}
We record a set of driving videos and extract one frame per second from the videos. Then, we manually label each frame with driving operations 1-accelerate, 2-brake, or 0-no operation and construct a small-scale \emph{Driving Control Dataset}. Some examples are presented in Figure \ref{fig: dataset}.

We verify the videos from the dataset against the following specifications:

$\Phi_1 = \lalways$ (pedestrian $\rightarrow$ brake),

$\Phi_2 = \lalways$ ((stop sign $\vee$ red light) $\rightarrow \lnext$ brake),

$\Phi_3 = \lalways$ ((red light $\land \lnext$ green light) $\rightarrow \lnext ( \leventually$ accelerate) ).

Our video search method can search for video clips that satisfy the specifications within a long video.
We select a five-minute driving video from the Driving Control Dataset, which is presented as 300 frames. We further divide them into ten half-minute chunks (each chunk has 30 frames) and verify each chunk against the three specifications. We present verification results in Figure \ref{fig: drive-spec}. By doing so, we can search for video clips that satisfy the specifications.

\section{Additional Results}
\subsection{Privacy Search on College Introductory Videos}
\label{sec: college}
\begin{figure}[t]
    \centering
    \includegraphics[width=0.59\linewidth]{figures/frames/harvard1.png}
    \includegraphics[width=0.395\linewidth]{figures/frames/harvard2.png}
    \includegraphics[width=0.59\linewidth]{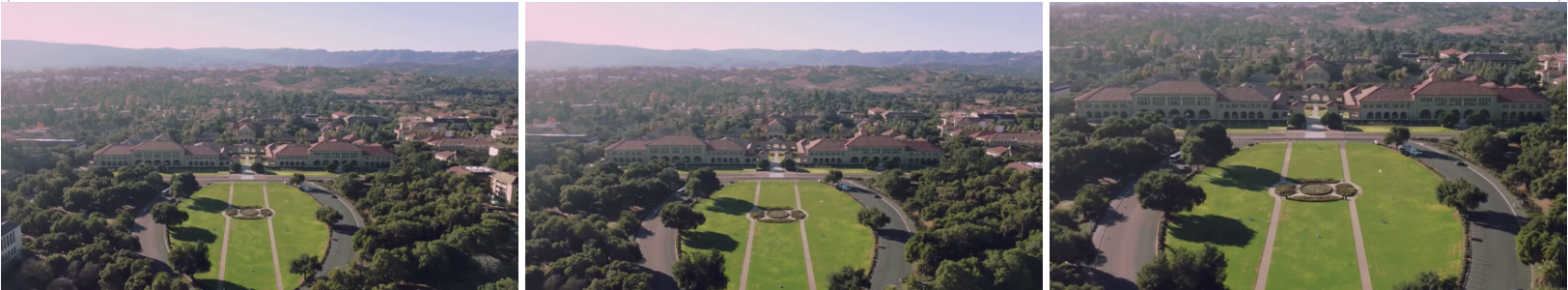}
    \includegraphics[width=0.395\linewidth]{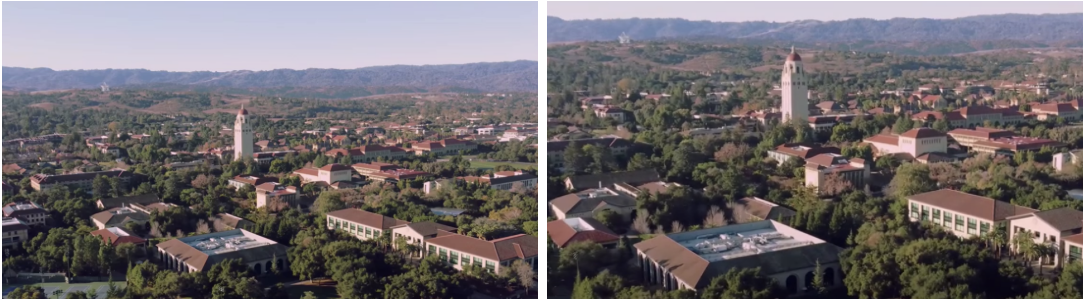}
    \includegraphics[width=0.59\linewidth]{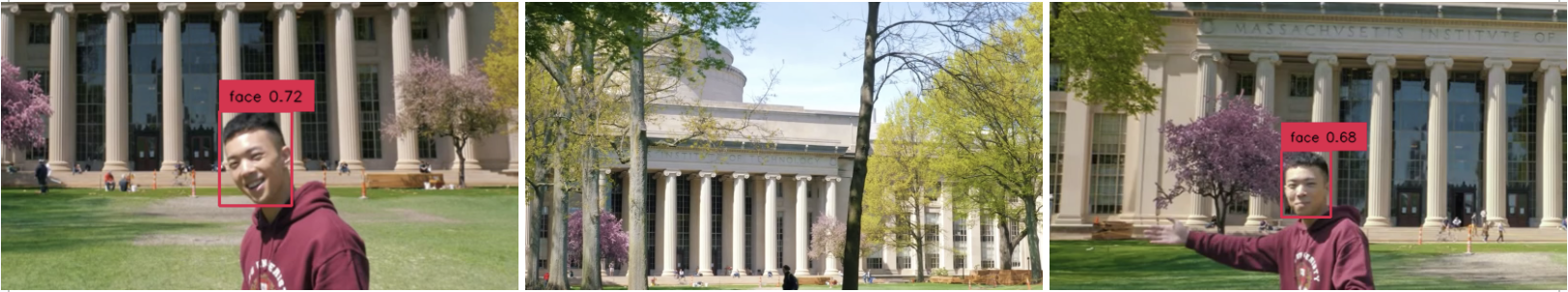}
    \includegraphics[width=0.395\linewidth]{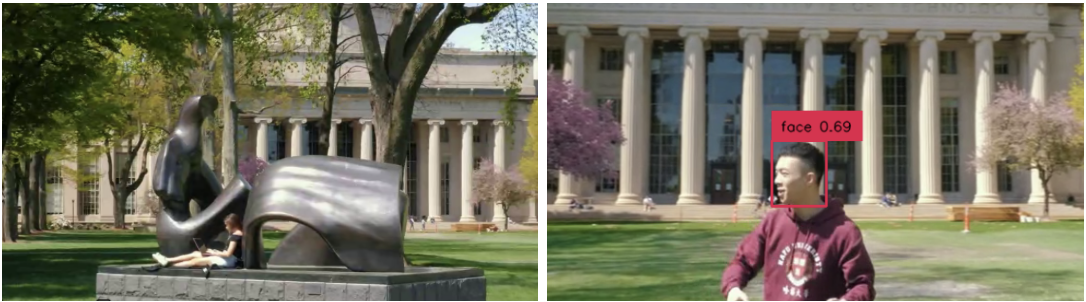}
    \caption{Object detection results on the frames from college introductory videos.}
    \label{fig: schools}
\end{figure}

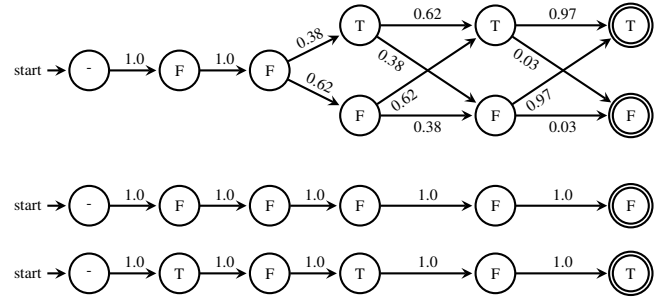
\begin{figure}
    \centering
    \begin{tikzpicture}[thick,scale=.6, node distance=2.2cm, every node/.style={transform shape}]
	\node[state, initial] (0) at (0, 0) {-};
	\node[state] (1) at (2, 0) {F};
    \node[state] (2) at (4, 0) {F};
    \node[state] (31) at (6, 1) {T};
	\node[state] (32) at (6, -1) {F};
	\node[state] (41) at (9, 1) {T};
    \node[state] (42) at (9, -1) {F};
    \node[state, accepting] (51) at (12, 1) {T};
    \node[state, accepting] (52) at (12, -1) {F};

	\draw[->, shorten >=1pt, sloped] (0) to[left] node[above, align=center] {1.0} (1);
    \draw[->, shorten >=1pt, sloped] (1) to[left] node[above, align=center] {1.0} (2);
    \draw[->, shorten >=1pt, sloped] (2) to[left] node[above, align=center] {0.38} (31);
    \draw[->, shorten >=1pt, sloped] (2) to[left] node[above, align=center] {0.62} (32);

    \draw[->, shorten >=1pt, sloped] (31) to[left] node[above, align=center] {0.62} (41);
    \draw[->, shorten >=1pt, sloped] (32) to[left] node[below, align=center, xshift=-8mm] {0.62} (41);
    \draw[->, shorten >=1pt, sloped] (31) to[left] node[below, align=center, xshift=-8mm] {0.38} (42);
    \draw[->, shorten >=1pt, sloped] (32) to[left] node[below, align=center] {0.38} (42);

    \draw[->, shorten >=1pt, sloped] (41) to[left] node[above, align=center] {0.97} (51);
    \draw[->, shorten >=1pt, sloped] (42) to[left] node[below, align=center, xshift=-8mm] {0.97} (51);
    \draw[->, shorten >=1pt, sloped] (41) to[left] node[below, align=center, xshift=-8mm] {0.03} (52);
    \draw[->, shorten >=1pt, sloped] (42) to[left] node[below, align=center] {0.03} (52);

    \node[state, initial] (s0) at (0, -3) {-};
	\node[state] (s1) at (2, -3) {F};
    \node[state] (s2) at (4, -3) {F};
    \node[state] (s3) at (6, -3) {F};
	\node[state] (s4) at (9, -3) {F};
    \node[state, accepting] (s5) at (12, -3) {F};

    \draw[->, shorten >=1pt, sloped] (s0) to[left] node[above, align=center] {1.0} (s1);
    \draw[->, shorten >=1pt, sloped] (s1) to[left] node[above, align=center] {1.0} (s2);
    \draw[->, shorten >=1pt, sloped] (s2) to[left] node[above, align=center] {1.0} (s3);
    \draw[->, shorten >=1pt, sloped] (s3) to[left] node[above, align=center] {1.0} (s4);
    \draw[->, shorten >=1pt, sloped] (s4) to[left] node[above, align=center] {1.0} (s5);

    \node[state, initial] (m0) at (0, -4.5) {-};
	\node[state] (m1) at (2, -4.5) {T};
    \node[state] (m2) at (4, -4.5) {F};
    \node[state] (m3) at (6, -4.5) {T};
	\node[state] (m4) at (9, -4.5) {F};
    \node[state, accepting] (m5) at (12, -4.5) {T};

    \draw[->, shorten >=1pt, sloped] (m0) to[left] node[above, align=center] {1.0} (m1);
    \draw[->, shorten >=1pt, sloped] (m1) to[left] node[above, align=center] {1.0} (m2);
    \draw[->, shorten >=1pt, sloped] (m2) to[left] node[above, align=center] {1.0} (m3);
    \draw[->, shorten >=1pt, sloped] (m3) to[left] node[above, align=center] {1.0} (m4);
    \draw[->, shorten >=1pt, sloped] (m4) to[left] node[above, align=center] {1.0} (m5);
 
\end{tikzpicture}
    \caption{Automata corresponding to the frames of the videos in Figure \ref{fig: schools}. `T' and `F' in each state indicate the state's label, either ``faces = True" or ``faces = False."}
    \label{fig: faces}
\end{figure}

We collect a set of college introductory videos from YouTube and verify each video against the following specifications:
\begin{center}
    $\Phi = \lalways \neg faces$.
\end{center}

We break each video into frames with a frequency of one frame per second. Next, we apply the Grounded-SAM object detection model to detect ``faces" in each frame and present the detection results in Figure \ref{fig: schools}.
Then, we follow Algorithm \ref{alg: frames2automata} to construct an automaton from the frames of each video, as presented in Figure \ref{fig: faces}.

We apply the Stormpy model checker to verify each video against the specification. From top to bottom, the automaton in Figure \ref{fig: faces} satisfies the specification with probabilities 0.007, 1.0, and 0, respectively. Among these three videos, the second video is the only one to be added to the search result.

\subsection{Verification on Traffic Recordings}
\label{sec: traffic}
\begin{figure}[t]
    \centering
    \includegraphics[width=\linewidth]{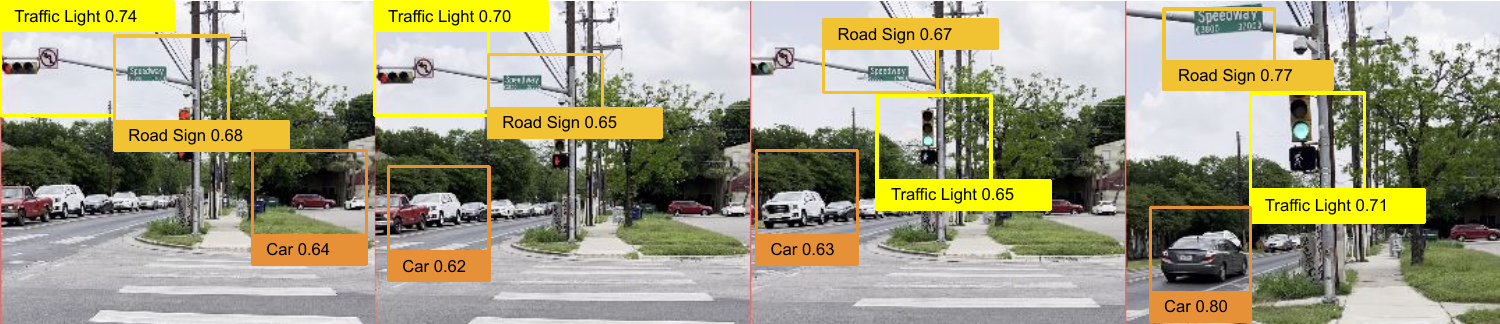}
    \caption{Object detection results on the frames of a manually recorded video from a road intersection.}
    \label{fig: traffic-frame}
\end{figure}

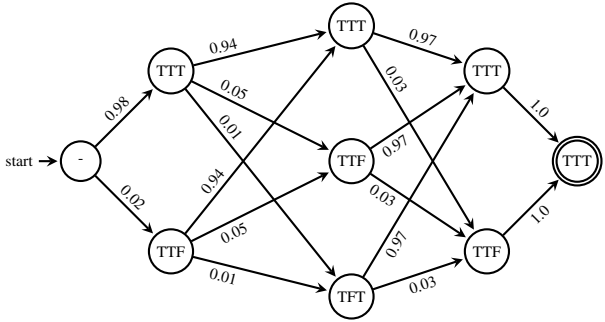
\begin{figure}[t]
    \centering
    \begin{tikzpicture}[thick,scale=.6, node distance=2.2cm, every node/.style={transform shape}]
    \node[state,initial] (0) at (-2, 0) {-};
    \node[state] (01) at (0, 2) {TTT};
    \node[state] (02) at (0, -2) {TTF};
    
    \node[state] (11) at (4, 3) {TTT};
    \node[state] (12) at (4, 0) {TTF};
    \node[state] (13) at (4, -3) {TFT};
    
    \node[state] (21) at (7, 2) {TTT};
    \node[state] (22) at (7, -2) {TTF};
    
    \node[state, accepting] (31) at (9, 0) {TTT};

    \draw[->, shorten >=1pt, sloped] (0) to[left] node[above, align=center] {0.98} (01);
    \draw[->, shorten >=1pt, sloped] (0) to[left] node[above, align=center] {0.02} (02);

    \draw[->, shorten >=1pt, sloped] (01) to[left] node[above, align=center, xshift=-8mm] {0.94} (11);
    \draw[->, shorten >=1pt, sloped] (01) to[left] node[above, align=center, xshift=-8mm] {0.05} (12);
    \draw[->, shorten >=1pt, sloped] (01) to[left] node[above, align=center, xshift=-14mm] {0.01} (13);

    \draw[->, shorten >=1pt, sloped] (02) to[left] node[above, align=center, xshift=-14mm] {0.94} (11);
    \draw[->, shorten >=1pt, sloped] (02) to[left] node[below, align=center, xshift=-8mm] {0.05} (12);
    \draw[->, shorten >=1pt, sloped] (02) to[left] node[below, align=center, xshift=-8mm] {0.01} (13);
    
    \draw[->, shorten >=1pt, sloped] (11) to[left] node[above, align=center] {0.97} (21);
    \draw[->, shorten >=1pt, sloped] (12) to[left] node[below, align=center, xshift=-8mm] {0.97} (21);
    \draw[->, shorten >=1pt, sloped] (13) to[left] node[below, align=center, xshift=-14mm] {0.97} (21);
    
    \draw[->, shorten >=1pt, sloped] (11) to[left] node[above, align=center, xshift=-14mm] {0.03} (22);
    \draw[->, shorten >=1pt, sloped] (12) to[left] node[below, align=center, xshift=-8mm] {0.03} (22);
    \draw[->, shorten >=1pt, sloped] (13) to[left] node[below, align=center] {0.03} (22);

    \draw[->, shorten >=1pt, sloped] (21) to[left] node[above, align=center] {1.0} (31);
    \draw[->, shorten >=1pt, sloped] (22) to[left] node[below, align=center] {1.0} (31);
    
\end{tikzpicture}
    \caption{Automaton constructed from the frames in Figure \ref{fig: traffic-frame}. `T' and `F' in each state indicate the ``True" and ``False" value of the corresponding label. The three values correspond to ``traffic light," ``road sign," and ``car," respectively. E.g., if a state is labeled as ``TTF," the label of this state is ``traffic light = True $\land$ road sign = True $\land$ car = False."}
    \label{fig: traffic}
\end{figure}

We collect traffic recording videos, extract frames in a frequency of one frame per second, and manually generate several rules regarding what may happen in a traffic recording:
\vspace{0.2cm}
\begin{lstlisting}[language=completion]
    1. Never reveal any road signs.
    2. At some point, simultaneously show a traffic light and a road sign.
    3. There are cars after seeing the traffic light.
\end{lstlisting}
\vspace{0.2cm}
We transform the rules above into the following \gls{tl} specifications:

$\Phi_1 = \lalways \neg$ road sign,
    
$\Phi_2 = \leventually$ traffic light $\land$ road sign,
    
$\Phi_3 = \text{ traffic light } \rightarrow \lnext$ car.

The specification $\Phi_3$ consists of a NEXT operator. Since we need to break the continuous timeline into discrete states to make the NEXT operation meaningful, we manually define NEXT as the next frame in this experiment.

While obtaining the specifications, we also get the atomic propositions $P = \{ \text{road sign, traffic light, car} \}$. We break each recording video into frames with a constant frequency of two frames per second. Then, we send each frame and the propositions as the input to the Grounded-SAM and obtain object detection results, as presented in Figure \ref{fig: traffic-frame}.

Given the object detection results with confidence scores and the estimated mapping function, we construct a probabilistic automaton following Algorithm \ref{alg: frames2automata} and present it in Figure \ref{fig: traffic}. Note that all the constructed automata have a set of input symbols $\Sigma = \{ \epsilon \}$.
Then, we use the probabilistic model checker to compute the probability of each specification being satisfied.

The probability of $\Phi_1$ being satisfied is 0, as all the trajectories have to reach the acceptance state while the ``road sign" is only evaluated to be true in the acceptance state. 
The probability of $\Phi_2$ being satisfied is 1, as all three propositions are evaluated to be true according to the detection result of the last frame.
The specification $\Phi_3$ is satisfied with a probability of 1. The proposition ``traffic light" is true according to the third frame in Figure \ref{fig: traffic-frame}, and the proposition ``road sign" is true according to the fourth frame. We define the NEXT operation as the next frame.

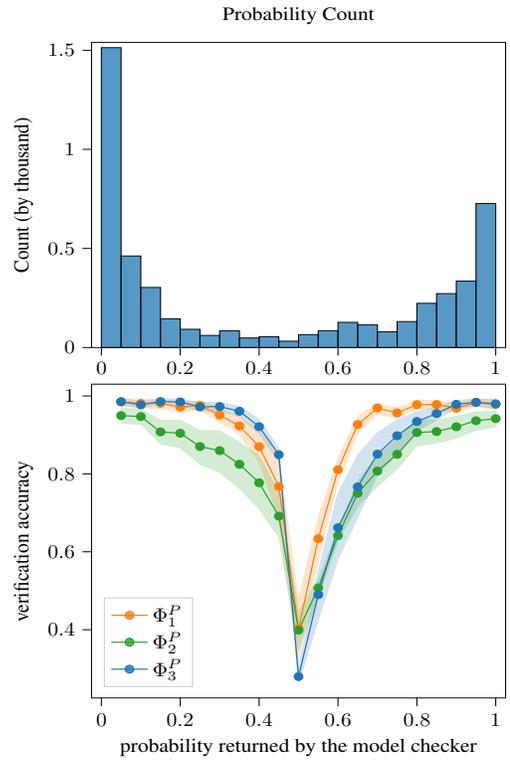
\begin{figure}[t]
    \centering
    \resizebox{0.8\linewidth}{0.6\linewidth}{% This file was created with tikzplotlib v0.10.1.
\begin{tikzpicture}

\definecolor{darkgray176}{RGB}{176,176,176}
\definecolor{steelblue31119180}{RGB}{31,119,180}

\begin{axis}[
tick align=outside,
tick pos=left,
title={Probability Count},
x grid style={darkgray176},
xmin=-0.025, xmax=1.025,
xtick style={color=black},
y grid style={darkgray176},
ylabel={Count (by thousand)},
ymin=0, ymax=1.54,
ytick style={color=black}
]
\draw[draw=black,fill=steelblue31119180,fill opacity=0.75] (axis cs:0,0) rectangle (axis cs:0.05,1.513);
\draw[draw=black,fill=steelblue31119180,fill opacity=0.75] (axis cs:0.05,0) rectangle (axis cs:0.1,0.462);
\draw[draw=black,fill=steelblue31119180,fill opacity=0.75] (axis cs:0.1,0) rectangle (axis cs:0.15,0.303);
\draw[draw=black,fill=steelblue31119180,fill opacity=0.75] (axis cs:0.15,0) rectangle (axis cs:0.2,0.145);
\draw[draw=black,fill=steelblue31119180,fill opacity=0.75] (axis cs:0.2,0) rectangle (axis cs:0.25,0.092);
\draw[draw=black,fill=steelblue31119180,fill opacity=0.75] (axis cs:0.25,0) rectangle (axis cs:0.3,0.061);
\draw[draw=black,fill=steelblue31119180,fill opacity=0.75] (axis cs:0.3,0) rectangle (axis cs:0.35,0.084);
\draw[draw=black,fill=steelblue31119180,fill opacity=0.75] (axis cs:0.35,0) rectangle (axis cs:0.4,0.048);
\draw[draw=black,fill=steelblue31119180,fill opacity=0.75] (axis cs:0.4,0) rectangle (axis cs:0.45,0.054);
\draw[draw=black,fill=steelblue31119180,fill opacity=0.75] (axis cs:0.45,0) rectangle (axis cs:0.5,0.032);
\draw[draw=black,fill=steelblue31119180,fill opacity=0.75] (axis cs:0.5,0) rectangle (axis cs:0.55,0.065);
\draw[draw=black,fill=steelblue31119180,fill opacity=0.75] (axis cs:0.55,0) rectangle (axis cs:0.6,0.085);
\draw[draw=black,fill=steelblue31119180,fill opacity=0.75] (axis cs:0.6,0) rectangle (axis cs:0.65,0.127);
\draw[draw=black,fill=steelblue31119180,fill opacity=0.75] (axis cs:0.65,0) rectangle (axis cs:0.7,0.114);
\draw[draw=black,fill=steelblue31119180,fill opacity=0.75] (axis cs:0.7,0) rectangle (axis cs:0.75,0.079);
\draw[draw=black,fill=steelblue31119180,fill opacity=0.75] (axis cs:0.75,0) rectangle (axis cs:0.8,0.131);
\draw[draw=black,fill=steelblue31119180,fill opacity=0.75] (axis cs:0.8,0) rectangle (axis cs:0.85,0.223);
\draw[draw=black,fill=steelblue31119180,fill opacity=0.75] (axis cs:0.85,0) rectangle (axis cs:0.9,0.272);
\draw[draw=black,fill=steelblue31119180,fill opacity=0.75] (axis cs:0.9,0) rectangle (axis cs:0.95,0.335);
\draw[draw=black,fill=steelblue31119180,fill opacity=0.75] (axis cs:0.95,0) rectangle (axis cs:1.0,0.727);
\end{axis}

\end{tikzpicture}}
    \resizebox{0.8\linewidth}{0.6\linewidth}{% This file was created with tikzplotlib v0.10.1.
\begin{tikzpicture}

\definecolor{darkgray176}{RGB}{176,176,176}
\definecolor{darkorange25512714}{RGB}{255,127,14}
\definecolor{forestgreen4416044}{RGB}{44,160,44}
\definecolor{steelblue31119180}{RGB}{31,119,180}

\definecolor{lightgray204}{RGB}{204,204,204}

\begin{axis}[
legend cell align={left},
legend style={
  fill opacity=0.8,
  draw opacity=1,
  text opacity=1,
  at={(0.03,0.32)},
  anchor=north west,
  draw=lightgray204
},
tick align=outside,
tick pos=left,
x grid style={darkgray176},
xlabel={probability returned by the model checker},
xmin=-0.025, xmax=1.025,
xtick style={color=black},
y grid style={darkgray176},
ylabel={verification accuracy},
ymin=0.22479297209645, ymax=1.0302808142003,
ytick style={color=black}
]
\path [draw=steelblue31119180, fill=steelblue31119180, opacity=0.2]
(axis cs:0.05,0.993350429760121)
--(axis cs:0.05,0.976617979287744)
--(axis cs:0.1,0.969263594736486)
--(axis cs:0.15,0.978027271193944)
--(axis cs:0.2,0.976314335570307)
--(axis cs:0.25,0.960988406263468)
--(axis cs:0.3,0.960632665572189)
--(axis cs:0.35,0.947382546932585)
--(axis cs:0.4,0.902759255130844)
--(axis cs:0.45,0.832249821185755)
--(axis cs:0.5,0.261406055828443)
--(axis cs:0.55,0.424664455706269)
--(axis cs:0.6,0.580591703623865)
--(axis cs:0.65,0.687381124082369)
--(axis cs:0.7,0.796684472531174)
--(axis cs:0.75,0.85364237542724)
--(axis cs:0.8,0.909325196360689)
--(axis cs:0.85,0.938065313508549)
--(axis cs:0.9,0.96761698752753)
--(axis cs:0.95,0.975126904241501)
--(axis cs:1,0.966782809676021)
--(axis cs:1,0.991855412641098)
--(axis cs:1,0.991855412641098)
--(axis cs:0.95,0.9921314217822)
--(axis cs:0.9,0.988338646432295)
--(axis cs:0.85,0.970426675517068)
--(axis cs:0.8,0.959489415241324)
--(axis cs:0.75,0.938784875375138)
--(axis cs:0.7,0.905618674507199)
--(axis cs:0.65,0.848694219993991)
--(axis cs:0.6,0.750256370632523)
--(axis cs:0.55,0.551751260158084)
--(axis cs:0.5,0.299357667993508)
--(axis cs:0.45,0.865727077704683)
--(axis cs:0.4,0.939747614356224)
--(axis cs:0.35,0.97421048160806)
--(axis cs:0.3,0.983596333696697)
--(axis cs:0.25,0.983313581862549)
--(axis cs:0.2,0.992019175333047)
--(axis cs:0.15,0.99298648249832)
--(axis cs:0.1,0.985343206196427)
--(axis cs:0.05,0.993350429760121)
--cycle;

\path [draw=darkorange25512714, fill=darkorange25512714, opacity=0.2]
(axis cs:0.05,0.993667730468309)
--(axis cs:0.05,0.973488314412182)
--(axis cs:0.1,0.969812638749674)
--(axis cs:0.15,0.967720067156759)
--(axis cs:0.2,0.958081225848864)
--(axis cs:0.25,0.963336783654311)
--(axis cs:0.3,0.933746996103516)
--(axis cs:0.35,0.901113955213554)
--(axis cs:0.4,0.829351516197667)
--(axis cs:0.45,0.685857830068809)
--(axis cs:0.5,0.320975277907558)
--(axis cs:0.55,0.576071809239786)
--(axis cs:0.6,0.779554939063565)
--(axis cs:0.65,0.904800436585993)
--(axis cs:0.7,0.954141989556484)
--(axis cs:0.75,0.943095490417097)
--(axis cs:0.8,0.966061731812494)
--(axis cs:0.85,0.965721668396947)
--(axis cs:0.9,0.956505093622547)
--(axis cs:0.95,0.972509795978929)
--(axis cs:1,0.965779821985908)
--(axis cs:1,0.98994349743235)
--(axis cs:1,0.98994349743235)
--(axis cs:0.95,0.992492188619416)
--(axis cs:0.9,0.980747412650193)
--(axis cs:0.85,0.988379771120408)
--(axis cs:0.8,0.98900430952332)
--(axis cs:0.75,0.970079301009551)
--(axis cs:0.7,0.981566452187333)
--(axis cs:0.65,0.947524008582569)
--(axis cs:0.6,0.840273348888017)
--(axis cs:0.55,0.688677127676306)
--(axis cs:0.5,0.476376735207849)
--(axis cs:0.45,0.842520584162754)
--(axis cs:0.4,0.91082929839521)
--(axis cs:0.35,0.94625570949949)
--(axis cs:0.3,0.96730089373206)
--(axis cs:0.25,0.987938288846416)
--(axis cs:0.2,0.98357007224519)
--(axis cs:0.15,0.992264701873688)
--(axis cs:0.1,0.991615340667903)
--(axis cs:0.05,0.993667730468309)
--cycle;

\path [draw=forestgreen4416044, fill=forestgreen4416044, opacity=0.2]
(axis cs:0.05,0.968201617645559)
--(axis cs:0.05,0.930215701488494)
--(axis cs:0.1,0.927116417250233)
--(axis cs:0.15,0.876611317184051)
--(axis cs:0.2,0.866925946372603)
--(axis cs:0.25,0.824547725966325)
--(axis cs:0.3,0.803622827592286)
--(axis cs:0.35,0.76271057147522)
--(axis cs:0.4,0.709616917883922)
--(axis cs:0.45,0.636829429944674)
--(axis cs:0.5,0.349607602296359)
--(axis cs:0.55,0.483223696930728)
--(axis cs:0.6,0.628059313345255)
--(axis cs:0.65,0.71605968941542)
--(axis cs:0.7,0.767591189572615)
--(axis cs:0.75,0.812203449550167)
--(axis cs:0.8,0.871782498522867)
--(axis cs:0.85,0.878875266250145)
--(axis cs:0.9,0.891917343258524)
--(axis cs:0.95,0.911600178431585)
--(axis cs:1,0.920895462123841)
--(axis cs:1,0.962091128541354)
--(axis cs:1,0.962091128541354)
--(axis cs:0.95,0.959066103418241)
--(axis cs:0.9,0.948067953107203)
--(axis cs:0.85,0.93786425043612)
--(axis cs:0.8,0.940058471390749)
--(axis cs:0.75,0.888738439989773)
--(axis cs:0.7,0.84562207021714)
--(axis cs:0.65,0.78143900694634)
--(axis cs:0.6,0.652869625450978)
--(axis cs:0.55,0.529636162970521)
--(axis cs:0.5,0.446624177944854)
--(axis cs:0.45,0.752270491493268)
--(axis cs:0.4,0.846438535633797)
--(axis cs:0.35,0.881224309696173)
--(axis cs:0.3,0.910658584086516)
--(axis cs:0.25,0.912090991114725)
--(axis cs:0.2,0.937623762913772)
--(axis cs:0.15,0.937755142102789)
--(axis cs:0.1,0.966243290804244)
--(axis cs:0.05,0.968201617645559)
--cycle;

\addplot [semithick, darkorange25512714, mark=*]
table {%
0.05 0.98447133816899
0.1 0.980890631452962
0.15 0.980902110815342
0.2 0.97115937114169
0.25 0.97544429301302
0.3 0.951086713209454
0.35 0.923112102230065
0.4 0.869853589389216
0.45 0.76744132748993
0.5 0.402757696398603
0.55 0.633625861341905
0.6 0.810564164035884
0.65 0.926812301648388
0.7 0.969048597256817
0.75 0.956511001010011
0.8 0.977672443734702
0.85 0.977916827312192
0.9 0.968811635554063
0.95 0.982932179718332
1 0.978583819639114
};
\addlegendentry{$\Phi_1^P$}

\addplot [semithick, forestgreen4416044, mark=*]
table {%
0.05 0.949667739449293
0.1 0.947214059883066
0.15 0.907716532479532
0.2 0.904347555490218
0.25 0.87022918243698
0.3 0.859649125080672
0.35 0.82463595872153
0.4 0.77713608319184
0.45 0.691355127568523
0.5 0.398826600703335
0.55 0.507316086648207
0.6 0.641579275065391
0.65 0.750235097271266
0.7 0.807239272965132
0.75 0.850131965730193
0.8 0.906146764778043
0.85 0.908738560311503
0.9 0.920900780235628
0.95 0.936337522698463
1 0.941781537503254
};
\addlegendentry{$\Phi_2^P$}

\addplot [semithick, steelblue31119180, mark=*]
table {%
0.05 0.985545121983048
0.1 0.977227177039543
0.15 0.985741541681543
0.2 0.984824314318939
0.25 0.972003433338648
0.3 0.972768059039854
0.35 0.96074254005685
0.4 0.921106372018068
0.45 0.849346755500218
0.5 0.279673493708256
0.55 0.489959618369621
0.6 0.661634668011808
0.65 0.766449520650113
0.7 0.850631152288143
0.75 0.897925326154069
0.8 0.934471760552729
0.85 0.954833140028798
0.9 0.978602178298043
0.95 0.983855671007872
1 0.979728604284026
};
\addlegendentry{$\Phi_3^P$}
\end{axis}

\end{tikzpicture}}
    \caption{The top figure shows the number count of verification results whose probabilities fall into each particular range. The bottom figure shows the accuracy of verification results within each range.}
    \label{fig: privacy-count}
\end{figure}
\subsection{Search over Privacy-Sensitive Videos}
In addition to the precision and recall in Definition \ref{def: metrics}, we present the video search accuracy and number count of videos whose probability of satisfying the specifications is between a particular range. We define \emph{video search accuracy} as the follow:
\label{sec: privacy}
\begin{definition}
    \label{def: acc}
    Let $\Phi$ be an \gls{tl} specification, $N$ be the number of videos whose verification probabilities returned by the model checker are between an interval $[c_1, c_2)$, $T_P$ be the number of videos whose verification probabilities are between $[c_1, c_2)$ that actually satisfy $\Phi$,
    we define \textsc{video search accuracy} $A_V$ as
    \begin{center}
        $A_V = 
        \begin{cases}
        \frac{N-T_P}{N} \text{ if } c_2 \le 0.5 \\
        \frac{T_P}{N} \text{ if } c_1 \ge 0.5.
        \end{cases}$
    \end{center}
\end{definition}
We present the number count and video search accuracies in Figure \ref{fig: privacy-count} and denote $N$ as the \emph{count} in the figure.

We observe that most of the videos have verification probabilities close to 0 or 1. The video search accuracy is above 95 percent when the verification probability is close to 0 or 1. And the accuracy reaches the lowest point (lower than 40 percent) when the probability is 0.5. Overall, our search method achieves above 90 percent accuracy over the privacy-annotated HMDB-51 dataset.

\subsection{Search over Autonomous Driving Videos}
\label{sec: autonomous}
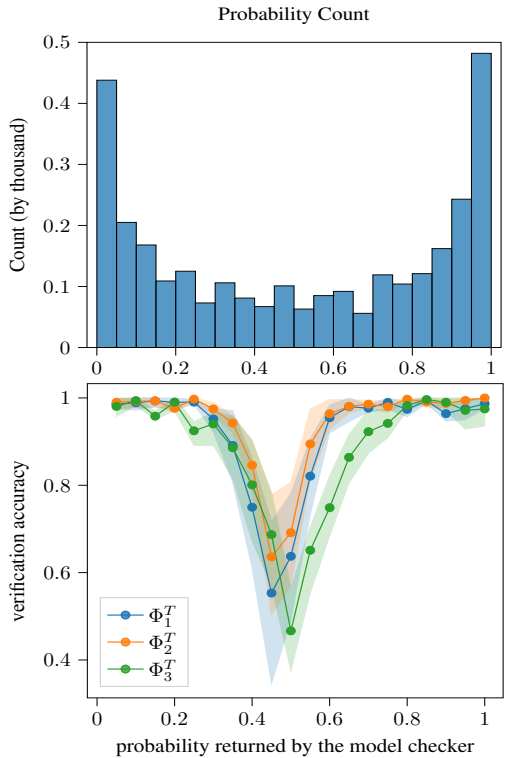
\begin{figure}[t]
    \centering
    \resizebox{0.8\linewidth}{0.6\linewidth}{% This file was created with tikzplotlib v0.10.1.
\begin{tikzpicture}

\definecolor{darkgray176}{RGB}{176,176,176}
\definecolor{steelblue31119180}{RGB}{31,119,180}

\begin{axis}[
tick align=outside,
tick pos=left,
title={Probability Count},
x grid style={darkgray176},
xmin=-0.025, xmax=1.025,
xtick style={color=black},
y grid style={darkgray176},
ylabel={Count (by thousand)},
ymin=0, ymax=0.5,
ytick style={color=black}
]
\draw[draw=black,fill=steelblue31119180,fill opacity=0.75] (axis cs:0,0) rectangle (axis cs:0.05,0.438);
\draw[draw=black,fill=steelblue31119180,fill opacity=0.75] (axis cs:0.05,0) rectangle (axis cs:0.1,0.205);
\draw[draw=black,fill=steelblue31119180,fill opacity=0.75] (axis cs:0.1,0) rectangle (axis cs:0.15,0.168);
\draw[draw=black,fill=steelblue31119180,fill opacity=0.75] (axis cs:0.15,0) rectangle (axis cs:0.2,0.109);
\draw[draw=black,fill=steelblue31119180,fill opacity=0.75] (axis cs:0.2,0) rectangle (axis cs:0.25,0.125);
\draw[draw=black,fill=steelblue31119180,fill opacity=0.75] (axis cs:0.25,0) rectangle (axis cs:0.3,0.073);
\draw[draw=black,fill=steelblue31119180,fill opacity=0.75] (axis cs:0.3,0) rectangle (axis cs:0.35,0.106);
\draw[draw=black,fill=steelblue31119180,fill opacity=0.75] (axis cs:0.35,0) rectangle (axis cs:0.4,0.081);
\draw[draw=black,fill=steelblue31119180,fill opacity=0.75] (axis cs:0.4,0) rectangle (axis cs:0.45,0.067);
\draw[draw=black,fill=steelblue31119180,fill opacity=0.75] (axis cs:0.45,0) rectangle (axis cs:0.5,0.101);
\draw[draw=black,fill=steelblue31119180,fill opacity=0.75] (axis cs:0.5,0) rectangle (axis cs:0.55,0.063);
\draw[draw=black,fill=steelblue31119180,fill opacity=0.75] (axis cs:0.55,0) rectangle (axis cs:0.6,0.085);
\draw[draw=black,fill=steelblue31119180,fill opacity=0.75] (axis cs:0.6,0) rectangle (axis cs:0.65,0.092);
\draw[draw=black,fill=steelblue31119180,fill opacity=0.75] (axis cs:0.65,0) rectangle (axis cs:0.7,0.056);
\draw[draw=black,fill=steelblue31119180,fill opacity=0.75] (axis cs:0.7,0) rectangle (axis cs:0.75,0.119);
\draw[draw=black,fill=steelblue31119180,fill opacity=0.75] (axis cs:0.75,0) rectangle (axis cs:0.8,0.104);
\draw[draw=black,fill=steelblue31119180,fill opacity=0.75] (axis cs:0.8,0) rectangle (axis cs:0.85,0.121);
\draw[draw=black,fill=steelblue31119180,fill opacity=0.75] (axis cs:0.85,0) rectangle (axis cs:0.9,0.162);
\draw[draw=black,fill=steelblue31119180,fill opacity=0.75] (axis cs:0.9,0) rectangle (axis cs:0.95,0.243);
\draw[draw=black,fill=steelblue31119180,fill opacity=0.75] (axis cs:0.95,0) rectangle (axis cs:1.0,0.482);
\end{axis}

\end{tikzpicture}}
    \resizebox{0.8\linewidth}{0.6\linewidth}{% This file was created with tikzplotlib v0.10.1.
\begin{tikzpicture}

\definecolor{darkgray176}{RGB}{176,176,176}
\definecolor{darkorange25512714}{RGB}{255,127,14}
\definecolor{forestgreen4416044}{RGB}{44,160,44}
\definecolor{steelblue31119180}{RGB}{31,119,180}

\definecolor{lightgray204}{RGB}{204,204,204}

\begin{axis}[
legend cell align={left},
legend style={
  fill opacity=0.8,
  draw opacity=1,
  text opacity=1,
  at={(0.03,0.32)},
  anchor=north west,
  draw=lightgray204
},
tick align=outside,
tick pos=left,
x grid style={darkgray176},
xlabel={probability returned by the model checker},
xmin=-0.025, xmax=1.0475,
xtick style={color=black},
y grid style={darkgray176},
ylabel={verification accuracy},
ymin=0.313260429057796, ymax=1.03270188433058,
ytick style={color=black}
]
\path [draw=steelblue31119180, fill=steelblue31119180, opacity=0.2]
(axis cs:0.05,0.996945860825289)
--(axis cs:0.05,0.974636354613159)
--(axis cs:0.1,0.972261197295331)
--(axis cs:0.15,0.979542704558064)
--(axis cs:0.2,0.974492589735314)
--(axis cs:0.25,0.9850349715919)
--(axis cs:0.3,0.922958819733535)
--(axis cs:0.35,0.806409256485894)
--(axis cs:0.4,0.611013307359541)
--(axis cs:0.45,0.345962313388378)
--(axis cs:0.5,0.491030442700137)
--(axis cs:0.55,0.733455636107666)
--(axis cs:0.6,0.919727379672894)
--(axis cs:0.65,0.945635626802159)
--(axis cs:0.7,0.96819496059591)
--(axis cs:0.75,0.973603702684791)
--(axis cs:0.8,0.95729439197826)
--(axis cs:0.85,0.987030590862907)
--(axis cs:0.9,0.946753249198977)
--(axis cs:0.95,0.950357853666542)
--(axis cs:1,0.97577894430588)
--(axis cs:1,0.998246925012311)
--(axis cs:1,0.998246925012311)
--(axis cs:0.95,0.998260697511306)
--(axis cs:0.9,0.98121239159791)
--(axis cs:0.85,1)
--(axis cs:0.8,0.990725855723409)
--(axis cs:0.75,1)
--(axis cs:0.7,0.984223816518194)
--(axis cs:0.65,1)
--(axis cs:0.6,0.983878724047964)
--(axis cs:0.55,0.906237409468082)
--(axis cs:0.5,0.780985971067399)
--(axis cs:0.45,0.716679405904122)
--(axis cs:0.4,0.861854058278338)
--(axis cs:0.35,0.970971395258727)
--(axis cs:0.3,0.975672299107973)
--(axis cs:0.25,0.996668121709011)
--(axis cs:0.2,1)
--(axis cs:0.15,1)
--(axis cs:0.1,1)
--(axis cs:0.05,0.996945860825289)
--cycle;

\path [draw=darkorange25512714, fill=darkorange25512714, opacity=0.2]
(axis cs:0.05,1)
--(axis cs:0.05,0.977477657471177)
--(axis cs:0.1,0.977466721467343)
--(axis cs:0.15,0.979944391389824)
--(axis cs:0.2,0.967096943092324)
--(axis cs:0.25,0.991497865288688)
--(axis cs:0.3,0.960085520779003)
--(axis cs:0.35,0.920370805388833)
--(axis cs:0.4,0.787471325743377)
--(axis cs:0.45,0.504270382421344)
--(axis cs:0.5,0.570004354490937)
--(axis cs:0.55,0.814602188033576)
--(axis cs:0.6,0.929222449842715)
--(axis cs:0.65,0.965390573551269)
--(axis cs:0.7,0.977460311580932)
--(axis cs:0.75,0.966068951027433)
--(axis cs:0.8,0.993100155075998)
--(axis cs:0.85,0.977765485285446)
--(axis cs:0.9,0.97746172504255)
--(axis cs:0.95,0.989373058641834)
--(axis cs:1,1)
--(axis cs:1,1)
--(axis cs:1,1)
--(axis cs:0.95,0.998299807512024)
--(axis cs:0.9,0.997236803483988)
--(axis cs:0.85,1)
--(axis cs:0.8,1)
--(axis cs:0.75,0.993427640849235)
--(axis cs:0.7,0.994633460059813)
--(axis cs:0.65,0.995922351414908)
--(axis cs:0.6,0.996058084609907)
--(axis cs:0.55,0.974534216334374)
--(axis cs:0.5,0.806878742382117)
--(axis cs:0.45,0.778117922391023)
--(axis cs:0.4,0.90216523548101)
--(axis cs:0.35,0.959518910296602)
--(axis cs:0.3,0.989399525813403)
--(axis cs:0.25,1)
--(axis cs:0.2,0.984233607159284)
--(axis cs:0.15,1)
--(axis cs:0.1,1)
--(axis cs:0.05,1)
--cycle;

\path [draw=forestgreen4416044, fill=forestgreen4416044, opacity=0.2]
(axis cs:0.05,0.998355785611009)
--(axis cs:0.05,0.957850503224462)
--(axis cs:0.1,0.982245310741297)
--(axis cs:0.15,0.950722558017341)
--(axis cs:0.2,0.970666978699248)
--(axis cs:0.25,0.891468924844153)
--(axis cs:0.3,0.8912437854496)
--(axis cs:0.35,0.815568808813993)
--(axis cs:0.4,0.67239368615099)
--(axis cs:0.45,0.570571987242811)
--(axis cs:0.5,0.375620068734824)
--(axis cs:0.55,0.55373212804477)
--(axis cs:0.6,0.684330713266778)
--(axis cs:0.65,0.80552892453881)
--(axis cs:0.7,0.873321537933967)
--(axis cs:0.75,0.909256426289006)
--(axis cs:0.8,0.970988379266457)
--(axis cs:0.85,0.988852855378028)
--(axis cs:0.9,0.970327343305148)
--(axis cs:0.95,0.928777899807093)
--(axis cs:1,0.93577462913815)
--(axis cs:1,0.997324767083776)
--(axis cs:1,0.997324767083776)
--(axis cs:0.95,1)
--(axis cs:0.9,1)
--(axis cs:0.85,1)
--(axis cs:0.8,0.994082025793797)
--(axis cs:0.75,0.973517636308111)
--(axis cs:0.7,0.972114910123317)
--(axis cs:0.65,0.91613205243375)
--(axis cs:0.6,0.821216036128858)
--(axis cs:0.55,0.704773960515971)
--(axis cs:0.5,0.563252140341499)
--(axis cs:0.45,0.784535258495787)
--(axis cs:0.4,0.902253755605563)
--(axis cs:0.35,0.958652446087497)
--(axis cs:0.3,0.967110287007214)
--(axis cs:0.25,0.945615814579194)
--(axis cs:0.2,1)
--(axis cs:0.15,0.969160672382538)
--(axis cs:0.1,1)
--(axis cs:0.05,0.998355785611009)
--cycle;

\addplot [semithick, steelblue31119180, mark=*]
table {%
0.05 0.985671824856185
0.1 0.988842741633301
0.15 0.993180901519355
0.2 0.989933002331936
0.25 0.990413101127564
0.3 0.951666153625208
0.35 0.891132859951746
0.4 0.749733397270965
0.45 0.553236916496756
0.5 0.63738074331841
0.55 0.82089516573679
0.6 0.954577442308443
0.65 0.979977304133951
0.7 0.976539146156413
0.75 0.989648658656735
0.8 0.973889542940845
0.85 0.993577640391707
0.9 0.963962629610269
0.95 0.975392633711913
1 0.98675156511345
};
\addlegendentry{$\Phi_1^T$}
\addplot [semithick, darkorange25512714, mark=*]
table {%
0.05 0.989894666213582
0.1 0.992488907155781
0.15 0.993314797129941
0.2 0.976229804151372
0.25 0.997165955096229
0.3 0.97490580932318
0.35 0.942603189448779
0.4 0.846143910591024
0.45 0.635999767952133
0.5 0.691412882248203
0.55 0.894568202183975
0.6 0.963825872293834
0.65 0.980656462483089
0.7 0.985698968348316
0.75 0.979897470058805
0.8 0.997066082232178
0.85 0.990752631023881
0.9 0.987570091478717
0.95 0.994185063837689
1 1
};
\addlegendentry{$\Phi_2^T$}
\addplot [semithick, forestgreen4416044, mark=*]
table {%
0.05 0.980554256369963
0.1 0.993770162741601
0.15 0.95818658541298
0.2 0.990222326233083
0.25 0.924748420958894
0.3 0.940450890362672
0.35 0.885850781234148
0.4 0.800881801582719
0.45 0.687197760645843
0.5 0.466745004338895
0.55 0.651235925814316
0.6 0.748731478080881
0.65 0.863861158873341
0.7 0.922810075481022
0.75 0.94176726572694
0.8 0.982590345943559
0.85 0.996284285126009
0.9 0.990109114435049
0.95 0.971487644388498
1 0.974843512548842
};
\addlegendentry{$\Phi_3^T$}
\end{axis}

\end{tikzpicture}}
    \caption{The top figure shows the number count of verification results whose probabilities fall into each particular range. The bottom figure shows the accuracy of verification results within each range.}
    \label{fig: traffic-count}
\end{figure}
We additionally present the number count and accuracy versus the verification probability over the object-annotated autonomous driving dataset. We observe that more verification results have probabilities close to 0 or 1, and fewer results have probabilities in the middle. The search accuracy for the videos with verification probability close to 0 or 1 is above 95 percent. However, the search accuracy is lower than 50 percent when the verification probability is approaching 0.5. In general, the overall search accuracy for the entire dataset is approximately 90 percent, which indicates the reliability of our video search method.

\end{document}